\documentclass[twoside,11pt]{article}

\usepackage{blindtext}

\usepackage{jmlr2e}
\usepackage{xcolor}
\usepackage{subcaption}
\usepackage{booktabs}
\usepackage{array}
\usepackage{multirow}

\usepackage{lastpage}
\jmlrheading{26}{2025}{1-\pageref{LastPage}}{12/19}{}{21-0000}{Guo, Zhu, Zhao, Zeng, Liu, Ma, Wang, Zhang}

\ShortHeadings{\texttt{MCITlib}: Multimodal Continual Instruction Tuning Library and Benchmark}{Guo, Zhu, Zhao, Zeng, Liu, Ma, Wang, Zhang}
\firstpageno{1}

\begin{document}

\title{\texttt{MCITlib}: Multimodal Continual Instruction Tuning Library and Benchmark}

\author{\name Haiyang Guo\textsuperscript{\rm 1,2} \email guohaiyang2023@ia.ac.cn \\
       \name Fei Zhu\textsuperscript{\rm 3} \email zhfei2018@gmail.com \\
       \name Hongbo Zhao\textsuperscript{\rm 2,4} \email zhaohongbo2022@ia.ac.cn \\
       \name Fanhu Zeng\textsuperscript{\rm 2,4} \email zengfanhu2022@ia.ac.cn \\
       \name Wenzhuo Liu\textsuperscript{\rm 2,4} \email liuwenzhuo2020@ia.ac.cn \\
       \name Shijie Ma\textsuperscript{\rm 2,4} \email mashijie2021@ia.ac.cn \\
       \name Da-Han Wang\textsuperscript{\rm 5} \email wangdh@xmut.edu.cn \\
       \name Xu-Yao Zhang\textsuperscript{\rm 1,2,4}\thanks{Xu-Yao Zhang is the corresponding author.} \email xyz@nlpr.ia.ac.cn \\
       \addr \textsuperscript{\rm 1}School of Advanced Interdisciplinary Sciences, University of Chinese Academy of Sciences, China \\
       \textsuperscript{\rm 2}State Key Laboratory of Multimodal Artificial Intelligence Systems, Institute of Automation, Chinese Academy of Sciences, China \\
       \textsuperscript{\rm 3}Centre for Artificial Intelligence and Robotics, Hong Kong Institute of Science and Innovation, Chinese Academy of Sciences, China \\
       \textsuperscript{\rm 4}School of Artificial Intelligence, University of Chinese Academy of Sciences, China \\
       \textsuperscript{\rm 5}Fujian Key Laboratory of Pattern Recognition and Image Understanding, School of Computer and Information Engineering, Xiamen University of Technology, China \\
       }
\editor{My editor}

\maketitle

\begin{abstract}
Continual learning enables AI systems to acquire new knowledge while retaining previously learned information. While traditional unimodal methods have made progress, the rise of Multimodal Large Language Models (MLLMs) brings new challenges in Multimodal Continual Learning (MCL), where models are expected to address both catastrophic forgetting and cross-modal coordination. To advance research in this area, we present \texttt{MCITlib}, a comprehensive library for Multimodal Continual Instruction Tuning. \texttt{MCITlib} currently implements 8 representative algorithms and conducts evaluations on 3 benchmarks under 2 backbone models. The library will be continuously updated to support future developments in MCL. The codebase is released at \url{https://github.com/Ghy0501/MCITlib}.
\end{abstract}


\section{Introduction}

Continual Learning~(CL), which aims to enable models to acquire and adapt knowledge continuously in a human-like manner, remains a fundamental challenge hindering the practical deployment of artificial intelligence systems in real-world scenarios. This difficulty primarily arises because models inevitably forget previously acquired knowledge when learning new information—a phenomenon known as \emph{catastrophic forgetting}~\citep{mccloskey1989catastrophic, french1999catastrophic, kirkpatrick2017overcoming}. Traditional continual learning research has primarily focused on unimodal tasks such as image classification or object detection, achieving remarkable progress~\citep{masana2022class, yuan2024survey, wang2024comprehensive}. However, the advent of Multimodal Large Language Models~(MLLMs)~\citep{yin2024survey} fundamentally broadens the scope of continual learning, introducing additional challenges in cross-modal alignment, knowledge integration, and modality-specific forgetting. In tandem with the burgeoning prominence of Multimodal Continual Learning~(MCL), there has been a surge in the development of various MCL algorithms and associated techniques~\citep{guo2025comprehensive}. Despite these advances, the lack of a unified and standardized platform for evaluating and comparing MCL methods hinders systematic progress in the field.

To bridge this gap, we propose \texttt{MCITlib}, a modular and continuously evolving codebase for Continual Instruction Tuning of MLLMs. \texttt{MCITlib} includes implementations of 8 representative Multimodal Continual Instruction Tuning (MCIT) algorithms, along with experiments on 3 continual instruction tuning benchmarks and 2 multimodal foundation models. The library is designed to be beginner-friendly and highly extensible, enabling contributors to seamlessly integrate new algorithms, thereby ensuring that \texttt{MCITlib} remains up-to-date and widely adopted. We believe that \texttt{MCITlib} can serve as a solid platform for continual learning researchers to investigate and develop methods in multimodal settings.

\section{Related Work}

As one of the long-standing research topics in machine learning, continual learning has given rise to a number of open-source platforms and code repositories. Most of these platforms are designed for continual learning settings in traditional computer vision tasks (\emph{e.g.,} image classification and segmentation), such as Avalanche~\citep{carta2023avalanche}, PILOT~\citep{sun2025pilot}, PyCIL~\citep{zhou2023pycil}, and CSSegmentation\footnote{~https://github.com/SegmentationBLWX/cssegmentation}. While these platforms have significantly advanced research in traditional continual learning tasks, they struggle to be directly applied to more complex models and tasks, especially with the rise of Large Language Models and MLLMs. There are also some open-source repositories for CL on Large Language Models, such as PyContinual\footnote{~https://github.com/ZixuanKe/PyContinual}, ContinualLM\footnote{~https://github.com/UIC-Liu-Lab/ContinualLM}, and \cite{zheng2024learn}. However, these platforms do not cover multimodal continual learning tasks and lack implementations of the latest continual learning algorithms.

CoIN~\citep{chen2024coin} is one of the latest MCIT projects; however, it provides only 4 continual learning algorithms, 2 of which are traditional methods, namely LwF~\citep{li2017learning} and EWC~\citep{kirkpatrick2017overcoming}. In contrast, our \texttt{MCITlib} provides implementations of 8 mainstream continual learning algorithms and has been extensively evaluated on 3 datasets and 2 multimodal foundation models. More importantly, \texttt{MCITlib} facilitates easy reproduction and extension, enabling users to integrate their own methods, datasets, and tasks with minimal effort.

\section{\texttt{MCITlib}: A User-Friendly Library for Multimodal Continual Learning}

\begin{figure}[t]
    \centering
    \includegraphics[width=0.98\linewidth]{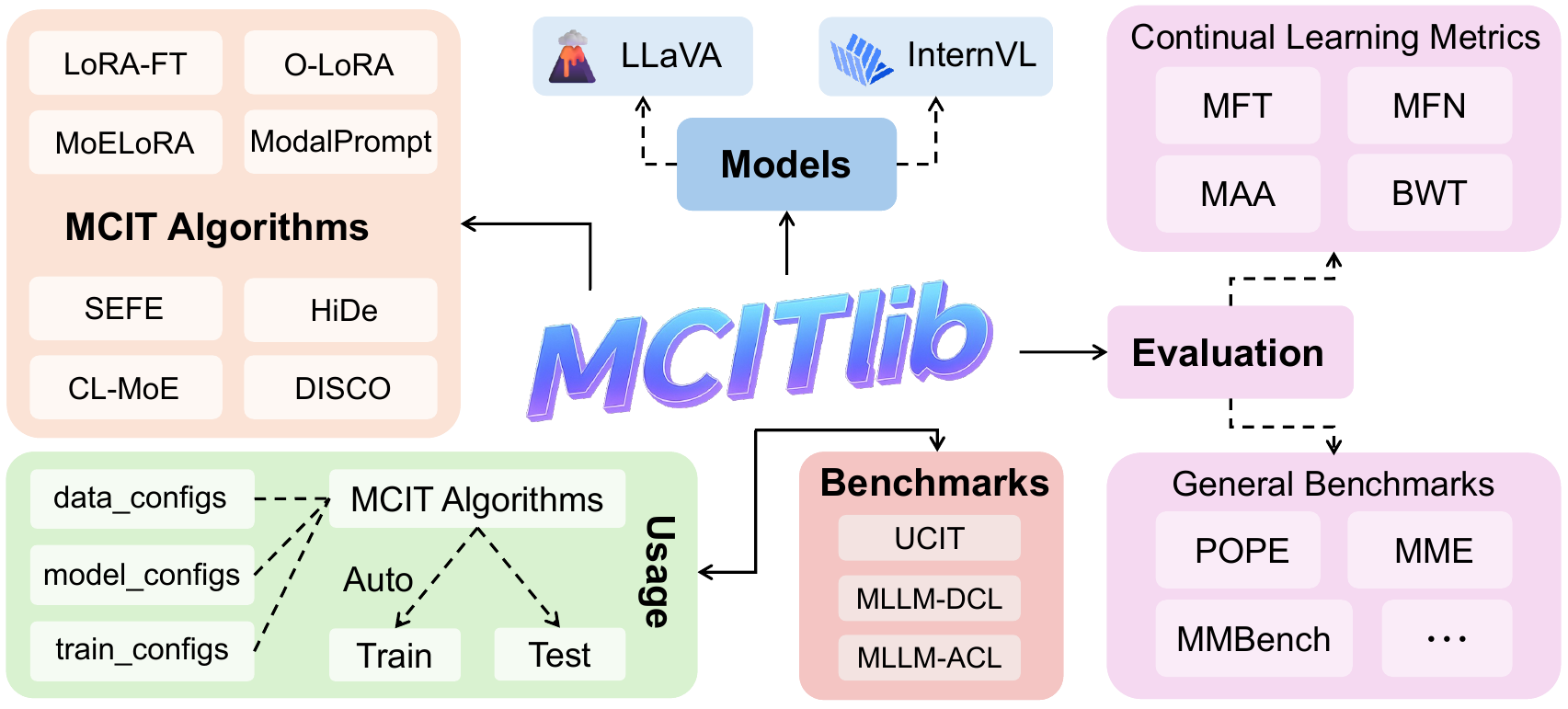}
    \caption{\texttt{MCITlib} main functionalities and modules.}
    \label{fig:figure1}
    \vspace{-15pt}
\end{figure}

\texttt{MCITlib} is implemented based on PyTorch~\citep{paszke2019pytorch} and consists of five main components: MCIT algorithms, models, benchmarks, evaluation, and usage. Figure~\ref{fig:figure1} illustrates the overall structure of \texttt{MCITlib}.

\noindent
\textbf{Algorithms and Models.} In \texttt{MCITlib}, we have implemented 8 MCIT algorithms, including LoRA-FT~\citep{hu2022lora}, O-LoRA~\citep{wang2023orthogonal}, MoELoRA~\citep{chen2024coin}, ModalPrompt~\citep{zeng2025modalprompt}, CL-MoE~\citep{huai2025cl}, HiDe-LLaVA~\citep{guo2025hide}, SEFE~\citep{chen2025sefe}, and DISCO~\citep{guo2025federated}. We adopt the commonly used LLaVA-1.5-7b~\citep{liu2024improved} and InternVL-Chat-7b~\citep{chen2024internvl} as the base models and employ Parameter-Efficient Fine-Tuning~(PEFT) strategies~\citep{hu2022lora, liu2023pre} for training. The training process follows the rehearsal-free continual learning setting~\citep{zhu2021prototype, zhu2025pass++}, where data from previous tasks is not reused during the training of new tasks. Implementation details are provided in Appendix~\ref{imp_detail}.

\noindent
\textbf{Benchmarks.} In selecting downstream tasks for continual learning with MLLMs, we adhere to the principle of avoiding information leakage~\citep{kim2023learnability}; that is, the downstream tasks should not overlap with the data used during the model’s pre-training or SFT stages, as such overlap could undermine the fairness and reliability of the evaluation. Accordingly, we selected the UCIT~\citep{guo2025hide}, MLLM-DCL, and MLLM-ACL~\citep{zhao2025mllm} benchmarks, which were identified as suitable downstream tasks for continual learning in MLLMs based on a comparison between the models’ zero-shot and fine-tuned performances. The detailed introduction of each benchmark is shown in the Appendix~\ref{mcitlib_details}.

\noindent
\textbf{Evaluation.} MCITlib evaluates along two axes: continual learning metrics and general-purpose benchmarks. For continual learning, following Chen et al. (2025), we report Mean Finetune Accuracy (MFT), Mean Final Accuracy (MFN), Mean Average Accuracy (MAA), and Backward Transfer (BWT); metric definitions are provided in Appendix~\ref{imp_detail}. For the latter, we note that for MLLMs with inherent generalization ability, it is essential to assess not only continual learning metrics but also the impact of different algorithms on the model’s original performance. Ideally, a method should prevent forgetting while enhancing the model’s overall capabilities. Hence, we use general multimodal benchmarks~\citep{fu2024mme} to evaluate this effect.

\noindent
\textbf{Usage.} \texttt{MCITlib} adopts a parameterized management framework: \textcolor{blue}{\texttt{data\_configs}} organizes benchmark paths, \textcolor{blue}{\texttt{model\_configs}} manages model weights, and \textcolor{blue}{\texttt{train\_configs}} defines training and inference parameters. This design facilitates flexible adjustments and efficient management by users. Once the relevant paths are configured, users can simply navigate to the directory of the selected algorithm and run  \textcolor{blue}{\texttt{sh scripts/MCITlib/Train/train\_XX.sh}} to automatically perform training and inference experiments.

\vspace{-10pt}
\begin{figure}[h]
    \centering
    \includegraphics[width=0.98\linewidth]{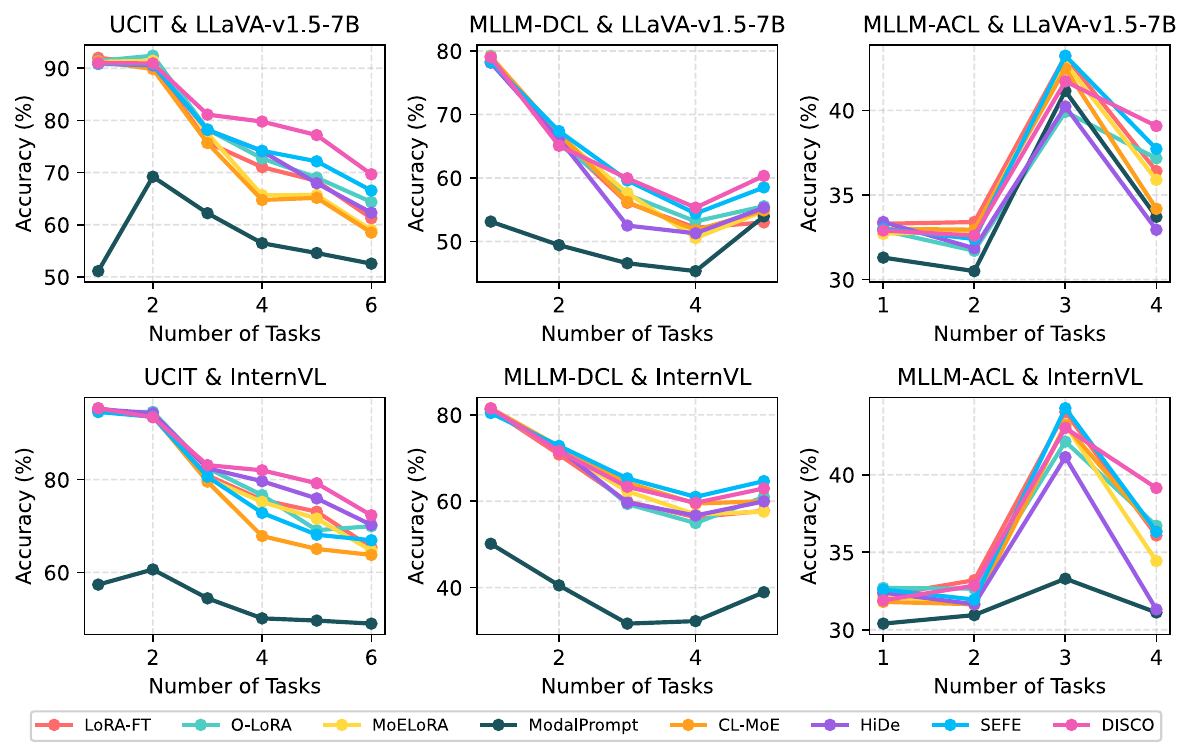}
    \caption{Performance curve of different methods under different settings.}
    \label{fig:figure2}
    \vspace{-25pt}
\end{figure}

\section{Experiments}

Figure~\ref{fig:figure2} shows accuracy curves for eight methods across three benchmarks and two backbones. Comprehensive CL metrics, general-purpose evaluations, and per-method result matrices are provided in Appendix~\ref{results_details}. Overall, current multimodal continual instruction methods partially mitigate forgetting on downstream tasks but transfer poorly to general-purpose benchmarks, often degrading the model’s original capabilities. Bridging this gap is a central direction for future work on continual learning for MLLMs.

\section{Conclusion}

In this paper, we introduce \texttt{MCITlib}, a comprehensive code library designed for continual instruction tuning of Multimodal Large Language Models. The library includes a collection of representative MCIT algorithms and carefully selected benchmarks that reduce information leakage and ensure fair comparisons. By providing unified implementations and evaluation protocols, \texttt{MCITlib} aims to accelerate research progress in Multimodal Continual Learning. 


\acks{This work was supported by the National Science and Technology Major Project (2022ZD-0116500), National Natural Science Foundation of China (62222609, 62320106010), CAS Project for Young Scientists in Basic Research (YSBR-083), Major Science and Technology Plan Project on the Future Industry Fields of Xiamen City (3502Z20241027), Unveiling and Leading Projects of Xiamen (3502Z20241011) and the InnoHK program. }


\bibliography{sample}

\newpage

\appendix
\section{Implementation Details}
\label{imp_detail}
\subsection{Training Details}

In Table~\ref{tab:train_config}, we summarize the training configurations for each method. For most training hyperparameters, we followed the original settings in LLaVA-1.5 and InternVL, adjusting parameters such as LoRA rank only according to the needs of different methods. Regarding the number of training epochs and the learning rate, we largely followed the settings in the UCIT and MLLM-CL papers, making trade-offs based on the performance of specific methods. As for parameters unique to a particular method, we set them according to the values in the original papers.

\begin{table*}[h]
    \centering
    \caption{Training configurations and PEFT settings for all methods. While MoELoRA follows a parameter-extension paradigm, it differs from other approaches in that it does not introduce task-specific modules. Accordingly, its settings are adjusted separately to ensure a fair comparison.}
    \resizebox{0.98\textwidth}{!}{
    \begin{tabular}{l|cccc|cccc}
        \toprule
        \multirow{2}{*}{\textbf{Methods}} & \multicolumn{4}{c|}{\textbf{LLaVA}} & \multicolumn{4}{c}{\textbf{InternVL}} \\ 
        \cmidrule{2-9}
        & LR & Epoch & PEFT & Parameter setting & LR & Epoch & PEFT & Parameter setting \\ \midrule
        \multicolumn{9}{c}{UCIT benchmark}  \\ \midrule
        LoRA-FT & 2e-4 & \{1,1,1,1,1,1\} & LoRA & rank = 16 & 1e-4 & \{1,1,1,1,1,1\} & LoRA & rank = 16 \\
        O-LoRA & 2e-4 & \{1,1,1,1,1,1\} & LoRA & rank = 96, expert num = 6 & 1e-4 & \{1,1,1,1,1,1\} & LoRA & rank = 96, expert num = 6 \\
        MoELoRA & 2e-4 & \{1,1,1,1,1,1\} & LoRA & rank = 18, expert num = 6 & 1e-4 & \{1,1,1,1,1,1\} & LoRA & rank = 18, expert num = 6 \\
        ModalPrompt & 2e-4 & \{1,1,1,1,1,1\} & Prompt & prefix\_len = 10, expert num = 6 & 1e-4 & \{1,1,1,1,1,1\} & Prompt & prefix\_len = 10, expert num = 6 \\
        CL-MoE & 2e-4 & \{1,1,1,1,1,1\} & LoRA & rank = 96, expert num = 6 & 1e-4 & \{1,1,1,1,1,1\} & LoRA & rank = 96, expert num = 6 \\
        HiDe & 2e-4 & \{1,1,1,1,1,1\} & LoRA & rank = 96, expert num = 6 & 1e-4 & \{1,1,1,1,1,1\} & LoRA & rank = 96, expert num = 6 \\
        SEFE & 2e-4 & \{1,1,1,1,1,1\} & LoRA & rank = 16 & 1e-4 & \{1,1,1,1,1,1\} & LoRA & rank = 16 \\
        DISCO & 2e-4 & \{1,1,1,1,1,1\} & LoRA & rank = 96, expert num = 6 & 1e-4 & \{1,1,1,1,1,1\} & LoRA & rank = 96, expert num = 6 \\ \midrule
        \multicolumn{9}{c}{MLLM-DCL benchmark}  \\ \midrule
        LoRA-FT & 2e-5 & \{1,3,1,2,1\} & LoRA & rank = 32 & 1e-4 & \{1,3,1,2,1\} & LoRA & rank = 32 \\
        O-LoRA & 2e-5 & \{1,3,1,2,1\} & LoRA & rank = 160, expert num = 5 & 1e-4 & \{1,3,1,2,1\} & LoRA & rank = 160, expert num = 5 \\
        MoELoRA & 2e-5 & \{1,3,1,2,1\} & LoRA & rank = 35, expert num = 5 & 1e-4 & \{1,3,1,2,1\} & LoRA & rank = 35, expert num = 5 \\
        ModalPrompt & 2e-4 & \{1,3,1,2,1\} & Prompt & prefix\_len = 20, expert num = 5 & 1e-4 & \{1,3,1,2,1\} & Prompt & prefix\_len = 20, expert num = 5 \\
        CL-MoE & 2e-5 & \{1,3,1,2,1\} & LoRA & rank = 160, expert num = 5 & 1e-4 & \{1,3,1,2,1\} & LoRA & rank = 160, expert num = 5 \\
        HiDe & 2e-5 & \{1,3,1,2,1\} & LoRA & rank = 160, expert num = 5 & 1e-4 & \{1,3,1,2,1\} & LoRA & rank = 160, expert num = 5 \\
        SEFE & 2e-5 & \{1,3,1,2,1\} & LoRA & rank = 32 & 1e-4 & \{1,3,1,2,1\} & LoRA & rank = 32 \\
        DISCO & 2e-5 & \{1,3,1,2,1\} & LoRA & rank = 160, expert num = 5 & 1e-4 & \{1,3,1,2,1\} & LoRA & rank = 160, expert num = 5 \\ \midrule
        \multicolumn{9}{c}{MLLM-ACL benchmark}  \\ \midrule
        LoRA-FT & 2e-5 & \{3,1,1,3\} & LoRA & rank = 32 & 1e-4 & \{3,1,1,3\} & LoRA & rank = 32 \\
        O-LoRA & 2e-5 & \{3,1,1,3\} & LoRA & rank = 128, expert num = 4 & 1e-4 & \{3,1,1,3\} & LoRA & rank = 128, expert num = 4 \\
        MoELoRA & 2e-5 & \{3,1,1,3\} & LoRA & rank = 32, expert num = 4 & 1e-4 & \{3,1,1,3\} & LoRA & rank = 32, expert num = 4 \\
        ModalPrompt & 2e-4 & \{3,1,1,3\} & Prompt & prefix\_len = 20, expert num = 4 & 1e-4 & \{3,1,1,3\} & Prompt & prefix\_len = 20, expert num = 4 \\
        CL-MoE & 2e-5 & \{3,1,1,3\} & LoRA & rank = 128, expert num = 4 & 1e-4 & \{3,1,1,3\} & LoRA & rank = 128, expert num = 4 \\
        HiDe & 2e-5 & \{3,1,1,3\} & LoRA & rank = 128, expert num = 4 & 1e-4 & \{3,1,1,3\} & LoRA & rank = 128, expert num = 4 \\
        SEFE & 2e-5 & \{3,1,1,3\} & LoRA & rank = 32 & 1e-4 & \{3,1,1,3\} & LoRA & rank = 32 \\
        DISCO & 2e-5 & \{3,1,1,3\} & LoRA & rank = 128, expert num = 4 & 1e-4 & \{3,1,1,3\} & LoRA & rank = 128, expert num = 4 \\
        \bottomrule
    \end{tabular}}
    \label{tab:train_config}
    \vspace{-10pt}
\end{table*}

\subsection{Evaluation Details}

\subsubsection{Continual Learning Metrics} 

Following the evaluation protocol in SEFE~\citep{chen2025sefe}, we assess the continual learning performance through a suite of four integrated metrics. These metrics are designed to provide a multi-faceted view of a model's ability to acquire new knowledge while preserving existing skills. The metrics are defined as follows:

\begin{itemize}
    \item \textbf{Mean Finetune Accuracy (MFT)}: The average accuracy on each task, evaluated immediately after its training concludes. This metric quantifies the model's learning capability on new tasks and serves as an empirical upper bound on performance, assuming no catastrophic forgetting.
    \item \textbf{Mean Final Accuracy (MFN)}: The average accuracy across all tasks, measured at the end of the entire training sequence. It reflects the overall knowledge retained by the model after learning all tasks.
    \item \textbf{Backward Transfer (BWT)}: Measures the influence of learning new tasks on the performance of previously learned tasks. It is calculated as the average difference between the final accuracy of each task and the accuracy obtained immediately after its initial training. A negative BWT value is a direct indicator of catastrophic forgetting.
    \item \textbf{Mean Average Accuracy (MAA)}: Provides a holistic performance measure throughout the learning process. It is computed by first calculating the average accuracy on all previously seen tasks after each task's training is complete and then averaging these values across all training steps.
\end{itemize}

\begin{figure}[h]
    \centering
    \includegraphics[width=0.49\linewidth]{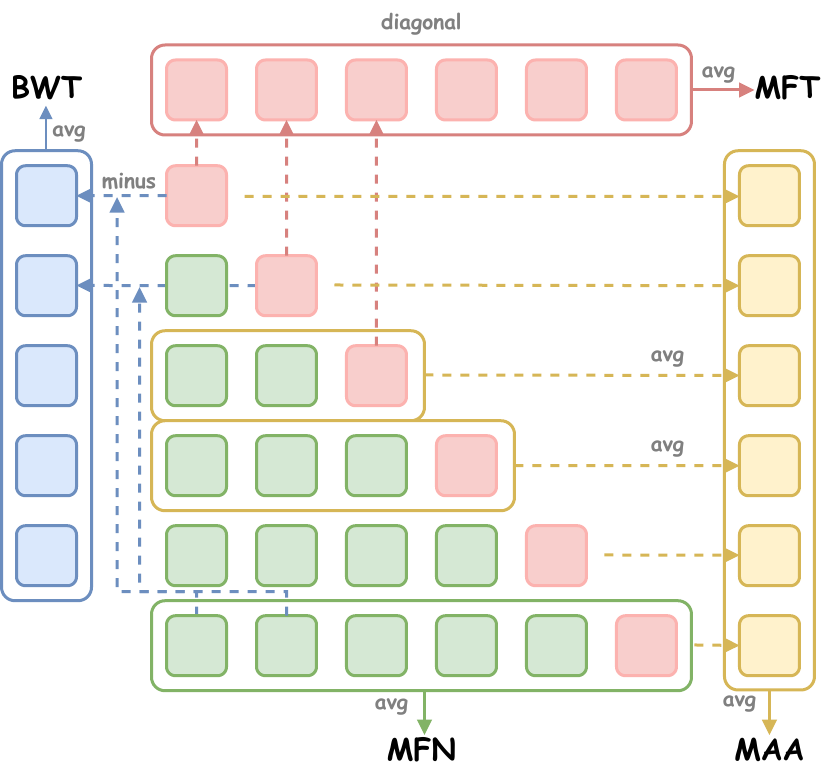}
    \caption{A conceptual illustration of the continual learning evaluation metrics.}
    \label{fig:figure3}
\end{figure}

A conceptual diagram illustrating these metrics is presented in Figure~\ref{fig:figure3}. We direct the reader to the original work by~\cite{chen2025sefe} for the rigorous mathematical formulations.

\subsubsection{General Benchmarks} 
Continual learning metrics primarily quantify how performance evolves over the sequence of learned tasks. However, for MLLMs that already exhibit broad generalization, it is insufficient to report CL metrics in isolation. Methods that score well on CL metrics can inadvertently erode general-purpose capability by over-fitting in the learned tasks. A desirable CL method sustains and preferably strengthens the model’s broad competence during sequential learning. Thus, we evaluate on four representative general-purpose benchmarks—POPE~\citep{li2023evaluating}, MME~\citep{fu2025mme}, MMBench~\citep{liu2024mmbench}, and SEED-Bench~\citep{li2023seed}. Below is a brief overview of each:

\begin{itemize}
    \item \textbf{POPE.} A benchmark for object hallucination and perception robustness in vision-language models. It measures whether the model invents non-existent objects under various prompts and contexts.
    \item \textbf{MME.} A comprehensive multimodal evaluation covering perception, knowledge, reasoning, and OCR-related skills, providing fine-grained sub-scores to diagnose capability gaps.
    \item \textbf{MMBench.} A broad, instruction-style benchmark with carefully curated, diverse question types spanning recognition, reasoning, commonsense, and multi-step inference; results are typically reported as accuracy under standardized prompting.
    \item \textbf{SEED-Bench.} A multi-dimensional assessment suite targeting generality and reliability, with tasks that stress instruction following, safety, factuality, and multimodal reasoning across varied domains.
\end{itemize}

We adhere to the official evaluation code wherever possible and integrate each benchmark into MCITlib’s automated workflow for one-click execution.

\section{MCIT Benchmarks Details}
\label{mcitlib_details}

\begin{figure}[h]
    \centering
    \includegraphics[width=0.98\linewidth]{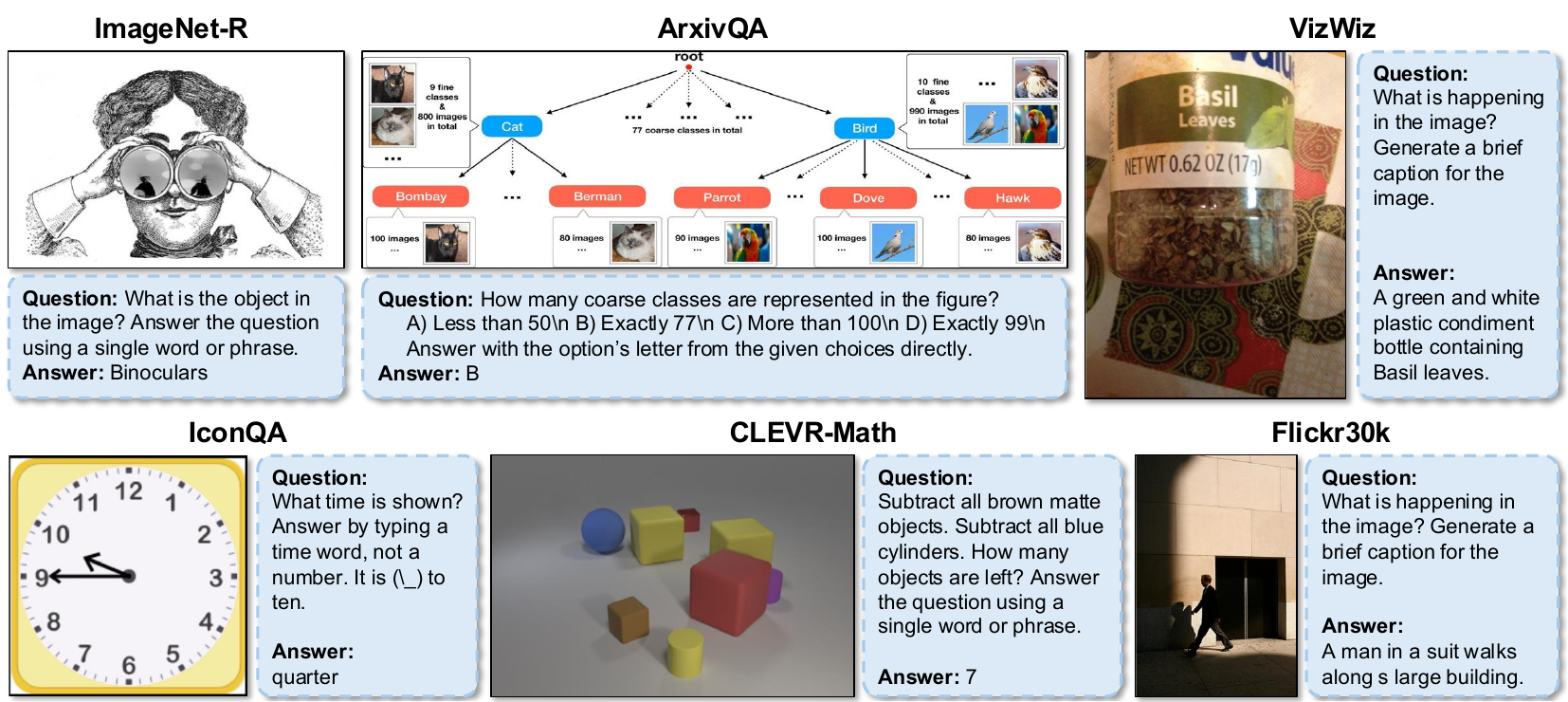}
    \caption{UCIT Benchmark Sample Visualization.}
    \label{fig:figure4}
\end{figure}

\begin{figure}[h]
    \centering
    \includegraphics[width=0.98\linewidth]{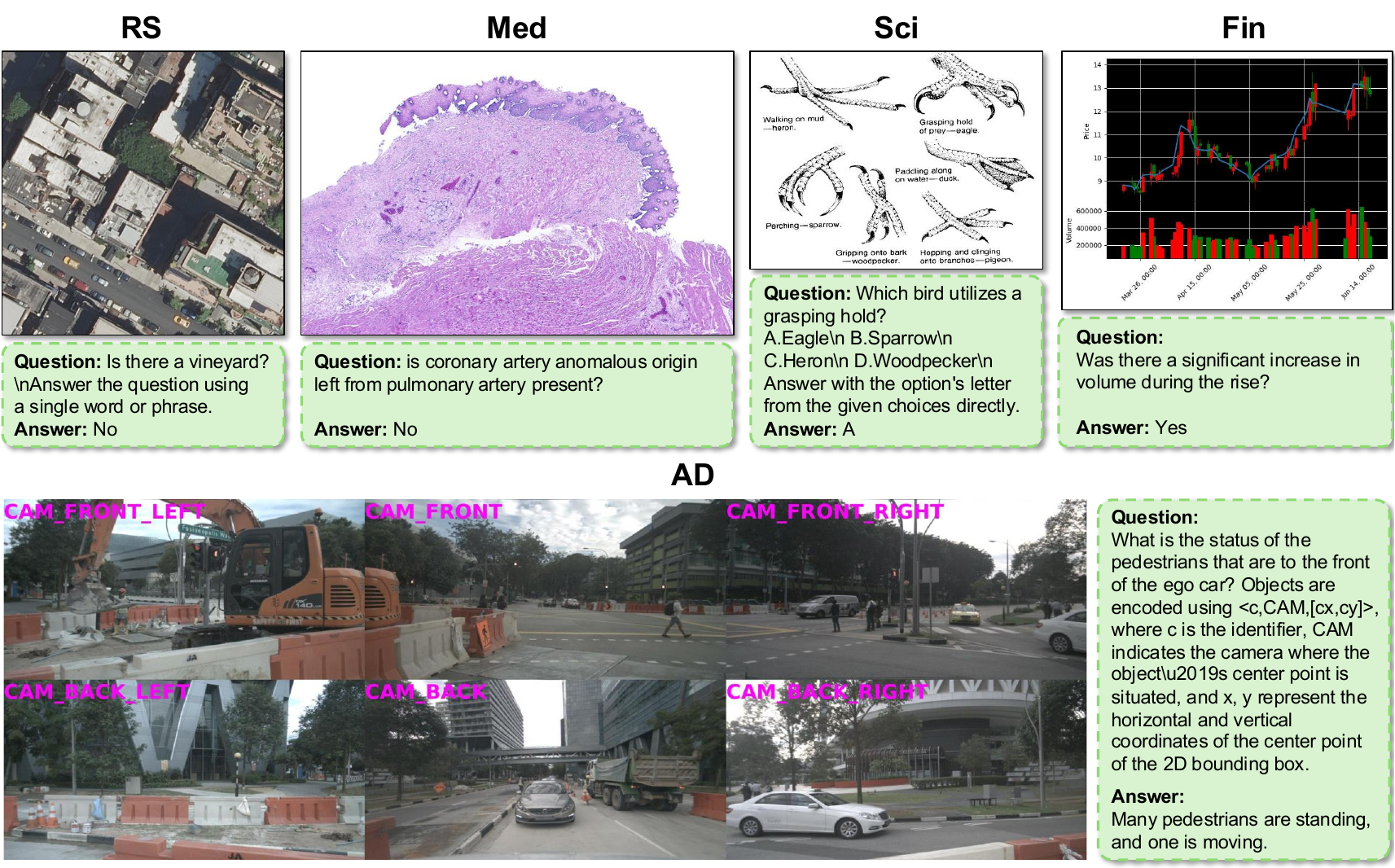}
    \caption{MLLM-DCL Benchmark Sample Visualization.}
    \label{fig:figure5}
\end{figure}

\begin{figure}[h]
    \centering
    \includegraphics[width=0.98\linewidth]{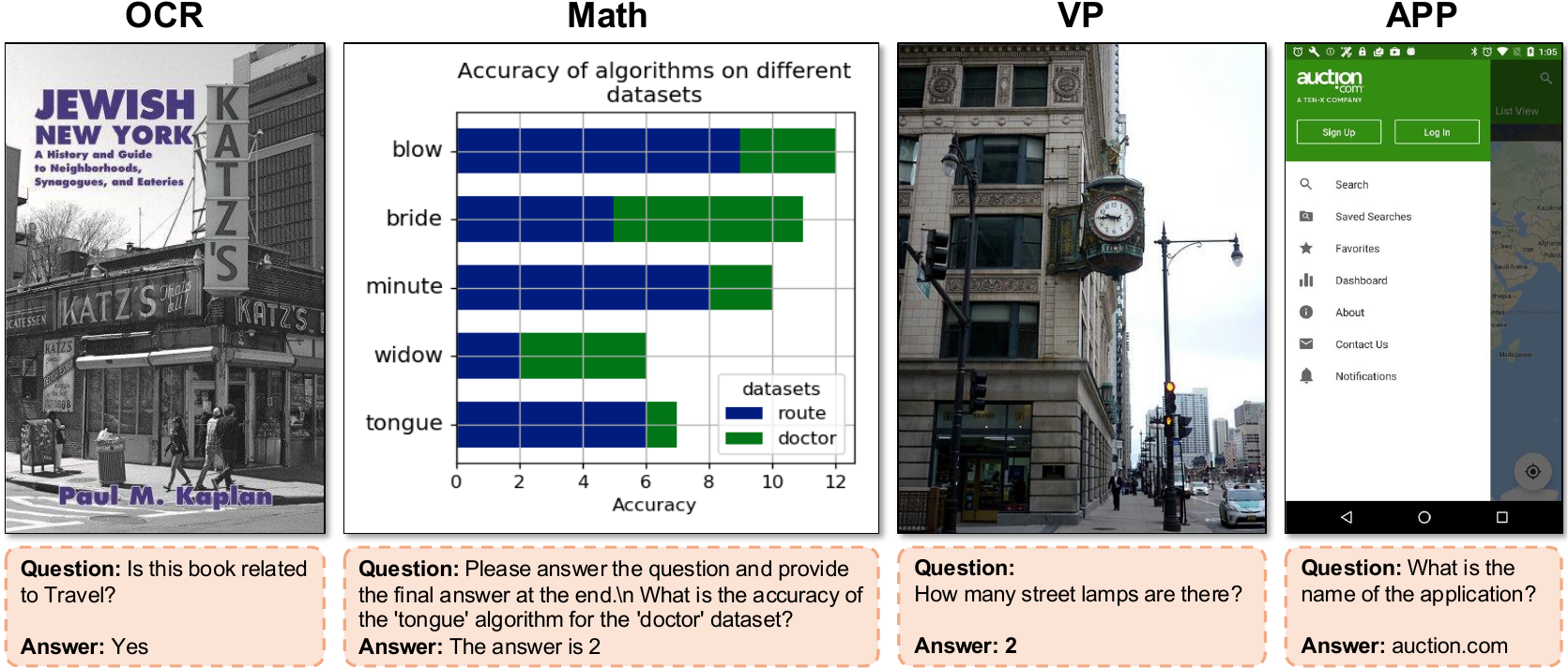}
    \caption{MLLM-ACL Benchmark Sample Visualization}
    \label{fig:figure6}
\end{figure}

In this section, we present statistics and visualizations for the benchmarks designed in \texttt{MCITlib}, as summarized in Table~\ref{Tab:static} and Figures \ref{fig:figure4}–\ref{fig:figure6}.

\begin{table}[h]
    \centering\small
    \caption{Statistics of the training datasets and test datasets for UCIT, MLLM-DCL and MLLM-ACL.}
    \renewcommand{\arraystretch}{1}
    \scalebox{0.9}{
    \begin{tabular}{lllcc}
    \toprule
Task & Train Dataset & Test Dataset & Train Number & Test Number  \\ \midrule 
\multicolumn{5}{c}{{UCIT}} \\ \midrule
ImgNet-R  &  ImageNet-R & ImageNet-R & 24k & 0.3k \\
ArxivQA  &  ArxivQA & ArxivQA & 40k & 0.3k \\
VizWiz  &  VizWiz & VizWiz & 40k & 0.3k \\
IconQA  &  IconQA & IconQA & 30k & 0.3k \\
CLEVR  &  CLEVR-Math & CLEVR-Math & 40k & 0.3k \\
Flickr30k  &  Flickr30k & Flickr30k & 40k & 0.3k \\
\midrule 
\multicolumn{5}{c}{{MLLM-DCL}} \\ \midrule
      RS  &  RSVQA & RSVQA & 60k & 10k \\
      Med  & PathVQA & PathVQA & 22.8k & 9.8k\\
      AD & DriveLM & DriveLM & 60k & 10k\\
      Sci &\begin{tabular}[c]{@{}l@{}}AI2D, SciVerse \\ MapQA, TQA\end{tabular}  & \begin{tabular}[c]{@{}l@{}}AI2D, SciVerse \\ MapQA, TQA\end{tabular} & \begin{tabular}[c]{@{}c@{}} 33.4k\\ (12.4k, 0.9k, 9.6k, 7.8k)\end{tabular}  & \begin{tabular}[c]{@{}c@{}} 8.2k \\ (3.1k, 0.2k, 2.4k, 1.9k)\end{tabular} \\
      Fin & StockQA &StockQA & 60k & 10k\\ \midrule
      \multicolumn{5}{c}{{MLLM-ACL}} \\ \midrule
      OCR & Monkey & OCRBench &128.1k & 1k \\
      Math & MathV360K, MAVIS & MathVista & 526.1k & 1k\\
      VP & CLEVR, TallyQA & CV-Bench &119.9k &0.8k\\
      GUI Agent & \begin{tabular}[c]{@{}l@{}}ScreenQA, MultiUI \\Screen2Words\end{tabular} & MMTBench &147.3k & 0.8k\\
      \bottomrule
    \end{tabular}
    }
    \label{Tab:static}
\end{table}

\section{Detailed Continual Learning Results}
\label{results_details}
\subsection{Continual Learning Metrics Results}

In this section, we report the continual learning performance for all methods across benchmarks and base models. In addition to the standard CL metrics introduced above, we also present each method’s test results on all tasks after training for the final task. Note that the original ModalPrompt paper recommends more training epochs for convergence. For fairness and runtime considerations, we report ModalPrompt using the same number of epochs as other methods, while ModalPrompt* denotes results with 10 epochs for all tasks. The detailed results are summarized in Tables~\ref{tab:table1}–\ref{tab:table2}. Results may vary across hardware and software environments. We recommend using the results obtained in the user's local setup as the primary reference.

\begin{table*}[t]
\centering
\caption{Comparison of different methods on LLaVA-1.5 and across multiple MCIT benchmarks. The best performance is shown in \textbf{bold}, and the second best is \underline{underlined}.}
\label{tab:table1}
\begin{subtable}[t]{0.98\linewidth}
\centering
\resizebox{\linewidth}{!}{
\begin{tabular}{>{\raggedright}p{2.0cm} *{11}{>{\centering\arraybackslash}p{1.2cm}}}
\toprule
\multicolumn{1}{>{\raggedright}p{2.5cm}|}{Method} & \multicolumn{1}{>{\raggedright}p{2.0cm}|}{Venue} & \multicolumn{1}{c}{ImgNet-R} & \multicolumn{1}{c}{ArxivQA} & \multicolumn{1}{c}{VizWiz} & \multicolumn{1}{c}{IconQA} & \multicolumn{1}{c}{CLEVR} & \multicolumn{1}{c|}{Flickr30k} & \multicolumn{1}{c}{MFT~($\uparrow$)} & \multicolumn{1}{c}{MFN~($\uparrow$)} & \multicolumn{1}{c}{MAA~($\uparrow$)} & \multicolumn{1}{c}{BWT~($\uparrow$)}\\
\midrule
\multicolumn{1}{>{\raggedright}p{2.5cm}|}{Zero-shot} & \multicolumn{1}{>{\raggedright}p{2.0cm}|}{-} & \multicolumn{1}{c}{16.27} & \multicolumn{1}{c}{53.73} & \multicolumn{1}{c}{38.39} & \multicolumn{1}{c}{19.20} & \multicolumn{1}{c}{20.63} & \multicolumn{1}{c|}{41.88} & \multicolumn{1}{c}{31.68} & \multicolumn{1}{c}{-} & \multicolumn{1}{c}{-} & \multicolumn{1}{c}{-}\\
\multicolumn{1}{>{\raggedright}p{2.5cm}|}{Individual} & \multicolumn{1}{>{\raggedright}p{2.0cm}|}{-} & \multicolumn{1}{c}{91.67} & \multicolumn{1}{c}{90.83} & \multicolumn{1}{c}{57.87} & \multicolumn{1}{c}{78.43} & \multicolumn{1}{c}{76.63} & \multicolumn{1}{c|}{61.72} & \multicolumn{1}{c}{76.19} & \multicolumn{1}{c}{-} & \multicolumn{1}{c}{-} & \multicolumn{1}{c}{-}\\
\midrule\midrule
\multicolumn{1}{>{\raggedright}p{2.5cm}|}{LoRA-FT} & \multicolumn{1}{>{\raggedright}p{2.0cm}|}{ICLR-22} & \multicolumn{1}{c}{58.03} & \multicolumn{1}{c}{77.63} & \multicolumn{1}{c}{44.39} & \multicolumn{1}{c}{67.40} & \multicolumn{1}{c}{61.77} & \multicolumn{1}{c|}{\textbf{58.22}} & \multicolumn{1}{c}{\textbf{76.89}} & \multicolumn{1}{c}{61.24} & \multicolumn{1}{c}{76.55} & \multicolumn{1}{c}{-18.78}\\
\bottomrule
\end{tabular}}
\vspace{3pt}
\caption{UCIT benchmark.}
\label{tab:table1_sub}
\end{subtable}

\begin{subtable}[t]{0.98\linewidth}
\centering
\resizebox{\linewidth}{!}{
\begin{tabular}{>{\raggedright}p{2.0cm} *{10}{>{\centering\arraybackslash}p{1.2cm}}}
\toprule
\multicolumn{1}{>{\raggedright}p{2.5cm}|}{Method} & \multicolumn{1}{>{\raggedright}p{2.0cm}|}{Venue} & \multicolumn{1}{c}{RS} & \multicolumn{1}{c}{Med} & \multicolumn{1}{c}{AD} & \multicolumn{1}{c}{Sci} & \multicolumn{1}{c|}{Fin} & \multicolumn{1}{c}{MFT~($\uparrow$)} & \multicolumn{1}{c}{MFN~($\uparrow$)} & \multicolumn{1}{c}{MAA~($\uparrow$)} & \multicolumn{1}{c}{BWT~($\uparrow$)}\\
\midrule
\multicolumn{1}{>{\raggedright}p{2.5cm}|}{Zero-shot} & \multicolumn{1}{>{\raggedright}p{2.0cm}|}{-} & \multicolumn{1}{c}{32.29} & \multicolumn{1}{c}{28.28} & \multicolumn{1}{c}{15.59} & \multicolumn{1}{c}{35.55} & \multicolumn{1}{c|}{62.56} & \multicolumn{1}{c}{34.85} & \multicolumn{1}{c}{-} & \multicolumn{1}{c}{-} & \multicolumn{1}{c}{-}\\
\multicolumn{1}{>{\raggedright}p{2.5cm}|}{Individual} & \multicolumn{1}{>{\raggedright}p{2.0cm}|}{-} & \multicolumn{1}{c}{78.15} & \multicolumn{1}{c}{58.20} & \multicolumn{1}{c}{52.77} & \multicolumn{1}{c}{49.32} & \multicolumn{1}{c|}{88.02} & \multicolumn{1}{c}{65.29} & \multicolumn{1}{c}{-} & \multicolumn{1}{c}{-} & \multicolumn{1}{c}{-}\\
\midrule\midrule
\multicolumn{1}{>{\raggedright}p{2.5cm}|}{LoRA-FT} & \multicolumn{1}{>{\raggedright}p{2.0cm}|}{ICLR-22} & \multicolumn{1}{c}{69.65} & \multicolumn{1}{c}{41.59} & \multicolumn{1}{c}{25.43} & \multicolumn{1}{c}{40.88} & \multicolumn{1}{c|}{87.45} & \multicolumn{1}{c}{64.98} & \multicolumn{1}{c}{53.00} & \multicolumn{1}{c}{61.13} & \multicolumn{1}{c}{-14.97}\\
\multicolumn{1}{>{\raggedright}p{2.5cm}|}{O-LoRA} & \multicolumn{1}{>{\raggedright}p{2.0cm}|}{EMNLP-23} & \multicolumn{1}{c}{74.64} & \multicolumn{1}{c}{44.42} & \multicolumn{1}{c}{30.02} & \multicolumn{1}{c}{41.47} & \multicolumn{1}{c|}{87.15} & \multicolumn{1}{c}{\underline{65.16}} & \multicolumn{1}{c}{55.54} & \multicolumn{1}{c}{62.12} & \multicolumn{1}{c}{-12.53}\\
\bottomrule
\end{tabular}}
\vspace{3pt}
\caption{MLLM-DCL benchmark.}
\label{tab:table2_sub}
\end{subtable}

\begin{subtable}[t]{0.98\linewidth}
\centering
\resizebox{\linewidth}{!}{
\begin{tabular}{>{\raggedright}p{2.0cm} *{9}{>{\centering\arraybackslash}p{1.2cm}}}
\toprule
\multicolumn{1}{>{\raggedright}p{2.5cm}|}{Method} & \multicolumn{1}{>{\raggedright}p{2.0cm}|}{Venue} & \multicolumn{1}{c}{OCR} & \multicolumn{1}{c}{Math} & \multicolumn{1}{c}{VP} & \multicolumn{1}{c|}{APP} & \multicolumn{1}{c}{MFT~($\uparrow$)} & \multicolumn{1}{c}{MFN~($\uparrow$)} & \multicolumn{1}{c}{MAA~($\uparrow$)} & \multicolumn{1}{c}{BWT~($\uparrow$)}\\
\midrule
\multicolumn{1}{>{\raggedright}p{2.5cm}|}{Zero-shot} & \multicolumn{1}{>{\raggedright}p{2.0cm}|}{-} & \multicolumn{1}{c}{31.20} & \multicolumn{1}{c}{30.20} & \multicolumn{1}{c}{60.79} & \multicolumn{1}{c|}{10.00} & \multicolumn{1}{c}{33.05} & \multicolumn{1}{c}{-} & \multicolumn{1}{c}{-} & \multicolumn{1}{c}{-}\\
\multicolumn{1}{>{\raggedright}p{2.5cm}|}{Individual} & \multicolumn{1}{>{\raggedright}p{2.0cm}|}{-} & \multicolumn{1}{c}{33.60} & \multicolumn{1}{c}{ 36.50} & \multicolumn{1}{c}{65.10} & \multicolumn{1}{c|}{\underline{32.50}} & \multicolumn{1}{c}{41.93} & \multicolumn{1}{c}{-} & \multicolumn{1}{c}{-} & \multicolumn{1}{c}{-}\\
\midrule\midrule
\multicolumn{1}{>{\raggedright}p{2.5cm}|}{LoRA-FT} & \multicolumn{1}{>{\raggedright}p{2.0cm}|}{ICLR-22} & \multicolumn{1}{c}{23.60} & \multicolumn{1}{c}{\textbf{33.70}} & \multicolumn{1}{c}{55.84} & \multicolumn{1}{c|}{\underline{32.50}} & \multicolumn{1}{c}{\textbf{41.28}} & \multicolumn{1}{c}{36.41} & \multicolumn{1}{c}{36.58} & \multicolumn{1}{c}{-6.49}\\
\multicolumn{1}{>{\raggedright}p{2.5cm}|}{O-LoRA} & \multicolumn{1}{>{\raggedright}p{2.0cm}|}{EMNLP-23} & \multicolumn{1}{c}{29.60} & \multicolumn{1}{c}{32.90} & \multicolumn{1}{c}{52.41} & \multicolumn{1}{c|}{\textbf{33.75}} & \multicolumn{1}{c}{39.72} & \multicolumn{1}{c}{37.17} & \multicolumn{1}{c}{35.42} & \multicolumn{1}{c}{-3.41}\\
\bottomrule
\end{tabular}}
\vspace{3pt}
\caption{MLLM-ACL benchmark.}
\label{tab:table3_sub}
\end{subtable}
\end{table*}

\begin{table*}[t]
\centering
\caption{Comparison of different methods on InternVL and across multiple MCIT benchmarks. The best performance is shown in \textbf{bold}, and the second best is \underline{underlined}.}
\label{tab:table2}
\begin{subtable}[t]{0.98\linewidth}
\centering
\resizebox{\linewidth}{!}{
\begin{tabular}{>{\raggedright}p{2.0cm} *{11}{>{\centering\arraybackslash}p{1.2cm}}}
\toprule
\multicolumn{1}{>{\raggedright}p{2.5cm}|}{Method} & \multicolumn{1}{>{\raggedright}p{2.0cm}|}{Venue} & \multicolumn{1}{c}{ImgNet-R} & \multicolumn{1}{c}{ArxivQA} & \multicolumn{1}{c}{VizWiz} & \multicolumn{1}{c}{IconQA} & \multicolumn{1}{c}{CLEVR} & \multicolumn{1}{c|}{Flickr30k} & \multicolumn{1}{c}{MFT~($\uparrow$)} & \multicolumn{1}{c}{MFN~($\uparrow$)} & \multicolumn{1}{c}{MAA~($\uparrow$)} & \multicolumn{1}{c}{BWT~($\uparrow$)}\\
\midrule
\multicolumn{1}{>{\raggedright}p{2.5cm}|}{Zero-shot} & \multicolumn{1}{>{\raggedright}p{2.0cm}|}{-} & \multicolumn{1}{c}{21.10} & \multicolumn{1}{c}{63.20} & \multicolumn{1}{c}{40.59} & \multicolumn{1}{c}{24.70} & \multicolumn{1}{c}{21.20} & \multicolumn{1}{c|}{44.67} & \multicolumn{1}{c}{35.91} & \multicolumn{1}{c}{-} & \multicolumn{1}{c}{-} & \multicolumn{1}{c}{-}\\
\multicolumn{1}{>{\raggedright}p{2.5cm}|}{Individual} & \multicolumn{1}{>{\raggedright}p{2.0cm}|}{-} & \multicolumn{1}{c}{91.67} & \multicolumn{1}{c}{90.83} & \multicolumn{1}{c}{57.87} & \multicolumn{1}{c}{78.43} & \multicolumn{1}{c}{76.63} & \multicolumn{1}{c|}{61.72} & \multicolumn{1}{c}{76.19} & \multicolumn{1}{c}{-} & \multicolumn{1}{c}{-} & \multicolumn{1}{c}{-}\\
\midrule\midrule
\multicolumn{1}{>{\raggedright}p{2.5cm}|}{LoRA-FT} & \multicolumn{1}{>{\raggedright}p{2.0cm}|}{ICLR-22} & \multicolumn{1}{c}{76.50} & \multicolumn{1}{c}{78.60} & \multicolumn{1}{c}{45.57} & \multicolumn{1}{c}{\underline{71.13}} & \multicolumn{1}{c}{61.63} & \multicolumn{1}{c|}{\textbf{58.82}} & \multicolumn{1}{c}{\underline{79.20}} & \multicolumn{1}{c}{65.38} & \multicolumn{1}{c}{80.66} & \multicolumn{1}{c}{-16.59}\\
\bottomrule
\end{tabular}}
\vspace{3pt}
\caption{UCIT benchmark.}
\label{tab:table4_sub}
\end{subtable}

\begin{subtable}[t]{0.98\linewidth}
\centering
\resizebox{\linewidth}{!}{
\begin{tabular}{>{\raggedright}p{2.0cm} *{10}{>{\centering\arraybackslash}p{1.2cm}}}
\toprule
\multicolumn{1}{>{\raggedright}p{2.5cm}|}{Method} & \multicolumn{1}{>{\raggedright}p{2.0cm}|}{Venue} & \multicolumn{1}{c}{RS} & \multicolumn{1}{c}{Med} & \multicolumn{1}{c}{AD} & \multicolumn{1}{c}{Sci} & \multicolumn{1}{c|}{Fin} & \multicolumn{1}{c}{MFT~($\uparrow$)} & \multicolumn{1}{c}{MFN~($\uparrow$)} & \multicolumn{1}{c}{MAA~($\uparrow$)} & \multicolumn{1}{c}{BWT~($\uparrow$)}\\
\midrule
\multicolumn{1}{>{\raggedright}p{2.5cm}|}{Zero-shot} & \multicolumn{1}{>{\raggedright}p{2.0cm}|}{-} & \multicolumn{1}{c}{31.16} & \multicolumn{1}{c}{29.81} & \multicolumn{1}{c}{14.06} & \multicolumn{1}{c}{33.93} & \multicolumn{1}{c|}{64.32} & \multicolumn{1}{c}{34.66} & \multicolumn{1}{c}{-} & \multicolumn{1}{c}{-} & \multicolumn{1}{c}{-}\\
\multicolumn{1}{>{\raggedright}p{2.5cm}|}{Individual} & \multicolumn{1}{>{\raggedright}p{2.0cm}|}{-} & \multicolumn{1}{c}{81.49} & \multicolumn{1}{c}{ 66.42} & \multicolumn{1}{c}{54.56} & \multicolumn{1}{c}{54.48} & \multicolumn{1}{c|}{\underline{91.24}} & \multicolumn{1}{c}{69.64} & \multicolumn{1}{c}{-} & \multicolumn{1}{c}{-} & \multicolumn{1}{c}{-}\\
\midrule\midrule
\multicolumn{1}{>{\raggedright}p{2.5cm}|}{LoRA-FT} & \multicolumn{1}{>{\raggedright}p{2.0cm}|}{ICLR-22} & \multicolumn{1}{c}{69.93} & \multicolumn{1}{c}{52.17} & \multicolumn{1}{c}{33.04} & \multicolumn{1}{c}{42.67} & \multicolumn{1}{c|}{91.07} & \multicolumn{1}{c}{69.06} & \multicolumn{1}{c}{57.78} & \multicolumn{1}{c}{65.22} & \multicolumn{1}{c}{-14.11}\\
\multicolumn{1}{>{\raggedright}p{2.5cm}|}{O-LoRA} & \multicolumn{1}{>{\raggedright}p{2.0cm}|}{EMNLP-23} & \multicolumn{1}{c}{74.48} & \multicolumn{1}{c}{54.16} & \multicolumn{1}{c}{39.60} & \multicolumn{1}{c}{\underline{48.30}} & \multicolumn{1}{c|}{88.54} & \multicolumn{1}{c}{65.51} & \multicolumn{1}{c}{61.02} & \multicolumn{1}{c}{65.83} & \multicolumn{1}{c}{\underline{-5.62}}\\
\bottomrule
\end{tabular}}
\vspace{3pt}
\caption{MLLM-DCL benchmark.}
\label{tab:table5_sub}
\end{subtable}

\begin{subtable}[t]{0.98\linewidth}
\centering
\resizebox{\linewidth}{!}{
\begin{tabular}{>{\raggedright}p{2.0cm} *{9}{>{\centering\arraybackslash}p{1.2cm}}}
\toprule
\multicolumn{1}{>{\raggedright}p{2.5cm}|}{Method} & \multicolumn{1}{>{\raggedright}p{2.0cm}|}{Venue} & \multicolumn{1}{c}{OCR} & \multicolumn{1}{c}{Math} & \multicolumn{1}{c}{VP} & \multicolumn{1}{c|}{APP} & \multicolumn{1}{c}{MFT~($\uparrow$)} & \multicolumn{1}{c}{MFN~($\uparrow$)} & \multicolumn{1}{c}{MAA~($\uparrow$)} & \multicolumn{1}{c}{BWT~($\uparrow$)}\\
\midrule
\multicolumn{1}{>{\raggedright}p{2.5cm}|}{Zero-shot} & \multicolumn{1}{>{\raggedright}p{2.0cm}|}{-} & \multicolumn{1}{c}{30.00} & \multicolumn{1}{c}{31.20} & \multicolumn{1}{c}{56.09} & \multicolumn{1}{c|}{2.50} & \multicolumn{1}{c}{29.95} & \multicolumn{1}{c}{-} & \multicolumn{1}{c}{-} & \multicolumn{1}{c}{-}\\
\multicolumn{1}{>{\raggedright}p{2.5cm}|}{Individual} & \multicolumn{1}{>{\raggedright}p{2.0cm}|}{-} & \multicolumn{1}{c}{32.20} & \multicolumn{1}{c}{33.40} & \multicolumn{1}{c}{67.77} & \multicolumn{1}{c|}{33.75} & \multicolumn{1}{c}{41.78} & \multicolumn{1}{c}{-} & \multicolumn{1}{c}{-} & \multicolumn{1}{c}{-}\\
\midrule\midrule
\multicolumn{1}{>{\raggedright}p{2.5cm}|}{LoRA-FT} & \multicolumn{1}{>{\raggedright}p{2.0cm}|}{ICLR-22} & \multicolumn{1}{c}{21.40} & \multicolumn{1}{c}{32.80} & \multicolumn{1}{c}{60.28} & \multicolumn{1}{c|}{29.86} & \multicolumn{1}{c}{40.85} & \multicolumn{1}{c}{36.09} & \multicolumn{1}{c}{\underline{36.38}} & \multicolumn{1}{c}{-6.35}\\
\multicolumn{1}{>{\raggedright}p{2.5cm}|}{O-LoRA} & \multicolumn{1}{>{\raggedright}p{2.0cm}|}{EMNLP-23} & \multicolumn{1}{c}{25.50} & \multicolumn{1}{c}{32.30} & \multicolumn{1}{c}{\underline{64.59}} & \multicolumn{1}{c|}{24.44} & \multicolumn{1}{c}{38.64} & \multicolumn{1}{c}{\underline{36.71}} & \multicolumn{1}{c}{36.05} & \multicolumn{1}{c}{-2.57}\\
\bottomrule
\end{tabular}}
\vspace{3pt}
\caption{MLLM-ACL benchmark.}
\label{tab:table6_sub}
\end{subtable}
\end{table*}

\subsection{General Benchmarks Results}

In this section, we evaluate the general benchmark performance using the final-task checkpoints obtained from MLLM-DCL training with different methods on two base models and compare them with the base models’ zero-shot performance. Results are shown in Figure~\ref{tab:table3}.

It can be seen that most continual instruction tuning methods for MLLMs diminish the model’s original capabilities. After sequentially learning downstream tasks, general-purpose performance consistently declines. This pattern suggests that existing approaches emphasize reducing forgetting on seen tasks while giving insufficient attention to preserving or improving overall competence. For MLLMs with strong inherent generalization, however, an effective continual learning method should both mitigate forgetting and maintain, or ideally enhance, general-purpose ability. We argue that this dual objective is a defining distinction between continual learning for MLLMs and traditional CL settings.

\begin{table*}[t]
\centering
\caption{Performance of downstream task weights learned on LLaVA-1.5/InternVL and MLLM-DCL with various methods, evaluated on general benchmarks and compared to the original models. $^\dagger$ denotes reproduced results. The best performance is shown in \textbf{bold}, and the second best is \underline{underlined}.}
\label{tab:table3}
\begin{subtable}[t]{0.98\linewidth}
\centering
\resizebox{\linewidth}{!}{%
\begin{tabular}{>{\raggedright\arraybackslash}p{2.6cm}|ccc|cc|c|ccc}
\toprule
\multirow{2}{*}{Method} & \multicolumn{3}{c|}{POPE} & \multicolumn{2}{c|}{MME} & \multirow{2}{*}{MMBench} & \multicolumn{3}{c}{SEED-Bench} \\
\cmidrule(lr){2-4} \cmidrule(lr){5-6} \cmidrule(lr){8-10}
& rand & pop & adv & per & cog & & all & img & vid \\
\midrule
LLaVA-1.5$^\dagger$ & 87.40 & \textbf{86.23} & \textbf{84.50} & \textbf{1483.50} & 323.57 & \textbf{65.90} & \textbf{61.50} & \textbf{67.31} & 39.33 \\
LoRA-FT & \textbf{88.52} & 85.87 & 82.37 & 1354.78 & 262.86 & 61.68 & 59.03 & 64.95 & 36.60 \\
O-LoRA & \underline{88.38} & 85.53 & 81.93 & 1384.33 & 305.36 & 62.20 & 58.75 & 64.72 & 36.12 \\
MoELoRA & 86.22 & 85.03 & 82.73 & 1383.36 & \underline{332.50} & 61.08 & 59.99 & 65.43 & 39.42 \\
ModalPrompt & 86.63 & 85.83 & \underline{83.60} & \underline{1473.79} & \underline{332.50} & \underline{64.43} & 60.11 & 65.35 & \underline{40.24} \\
CL-MoE & 86.84 & 85.57 & 83.03 & 1427.84 & 293.93 & 62.03 & 58.25 & 63.75 & 37.40 \\
HiDE & 88.14 & \underline{86.20} & 82.87 & 1349.76 & 295.71 & 55.07 & 59.56 & 64.95 & 39.15 \\
SEFE & 87.56 & 85.97 & 83.10 & 1461.71 & \textbf{337.50} & 62.46 & 60.10 & 65.76 & 38.67 \\
DISCO & 83.23 & 83.43 & 82.10 & 1363.59 & 300.36 & 63.75 & \underline{61.02} & \underline{66.42} & \textbf{40.56} \\
\bottomrule
\end{tabular}%
}
\vspace{3pt}
\caption{LLaVA-1.5 \& MLLM-DCL.}
\label{tab:table7_sub}
\end{subtable}
\hfill
\begin{subtable}[t]{0.98\linewidth}
\centering
\resizebox{\linewidth}{!}{%
\begin{tabular}{>{\raggedright\arraybackslash}p{2.6cm}|ccc|cc|c|ccc}
\toprule
\multirow{2}{*}{Method} & \multicolumn{3}{c|}{POPE} & \multicolumn{2}{c|}{MME} & \multirow{2}{*}{MMBench} & \multicolumn{3}{c}{SEED-Bench} \\
\cmidrule(lr){2-4} \cmidrule(lr){5-6} \cmidrule(lr){8-10}
& rand & pop & adv & per & cog & & all & img & vid \\
\midrule
InternVL$^\dagger$ & \underline{85.67} & \textbf{84.97} & \textbf{83.07} & \textbf{1477.26} & \underline{358.93} & \textbf{65.12} & \textbf{60.28} & \textbf{65.49} & \textbf{40.56} \\
LoRA-FT & 83.16 & 82.73 & \underline{79.17} & 1222.88 & 343.57 & 60.54 & 51.97 & 56.90 & 33.30 \\
O-LoRA & 81.75 & 77.93 & 75.13 & 1079.55 & 239.64 & 46.13 & 47.01 & 50.53 & 33.67 \\
MoELoRA & 85.43 & 82.83 & 79.07 & 1241.42 & 344.29 & 57.13 & 52.82 & 57.33 & 35.75 \\
ModalPrompt & 76.84 & 77.43 & 76.30 & \underline{1380.82} & 318.21 & \underline{62.80} & \underline{57.04} & \underline{62.05} & \underline{38.04} \\
CL-MoE & 83.85 & 82.03 & 78.70 & 1255.60 & \textbf{382.50} & 53.18 & 49.97 & 54.40 & 33.19 \\
HiDE & \textbf{85.84} & \underline{83.07} & 78.27 & 1122.21 & 261.79 & 45.96 & 48.54 & 52.46 & 33.67 \\
SEFE & 83.33 & 82.50 & 78.53 & 1244.08 & 336.07 & 51.98 & 49.38 & 53.44 & 33.99 \\
DISCO & 76.17 & 75.88 & 75.20 & 1116.63 & 312.86 & 57.22 & 53.37 & 57.90 & 36.23 \\
\bottomrule
\end{tabular}%
}
\vspace{3pt}
\caption{InternVL \& MLLM-DCL.}
\label{tab:table8_sub}
\end{subtable}
\end{table*}

\subsection{Detailed Result Matrixs}

In Tables~\ref{tab:table4}–\ref{tab:table9}, we report the final accuracy matrices for all methods under all settings. Results may vary across hardware and software environments. We recommend using the results obtained in the user's local setup as the primary reference.

\begin{table*}[t]
\centering
\caption{Result matrices for different methods on the UCIT benchmark and LLaVA‑1.5.}
\label{tab:table4}
\begin{subtable}[t]{0.49\linewidth}
\centering
\resizebox{\linewidth}{!}{
\begin{tabular}{lcccccc}
\toprule
\textbf{LoRA-FT} & ImgNet-R & ArxivQA & VizWiz & IconQA & CLEVR & Flickr30k \\
\midrule
ImgNet-R & 92.00 & & & & & \\
ArxivQA & 90.63 & 91.03 & & & & \\
VizWiz & 74.00 & 91.40 & 61.90 & & & \\
IconQA & 72.73 & 78.53 & 53.03 & 79.87 & & \\
CLEVR & 68.97 & 77.60 & 49.43 & 67.67 & 78.33 & \\
Flickr30k & 58.03 & 77.63 & 44.39 & 67.40 & 61.77 & 58.22 \\
\bottomrule
\end{tabular}}
\end{subtable}\hfill
\begin{subtable}[t]{0.49\linewidth}
\centering
\resizebox{\linewidth}{!}{
\begin{tabular}{lcccccc}
\toprule
\textbf{O-LoRA} & ImgNet-R & ArxivQA & VizWiz & IconQA & CLEVR & Flickr30k \\
\midrule
ImgNet-R & 91.40 & & & & & \\
ArxivQA & 90.07 & 94.73 & & & & \\
VizWiz & 81.50 & 91.70 & 61.59 & & & \\
IconQA & 82.40 & 77.73 & 54.06 & 76.63 & & \\
CLEVR & 81.07 & 77.87 & 50.62 & 62.00 & 73.53 & \\
Flickr30k & 77.50 & 78.07 & 44.50 & 63.13 & 64.73 & 58.16 \\
\bottomrule
\end{tabular}}
\end{subtable}
\begin{subtable}[t]{0.49\linewidth}
    \centering
    \resizebox{\linewidth}{!}{
    \begin{tabular}{lcccccc}
    \toprule
        \textbf{MoELoRA} & ImgNet-R & ArxivQA & VizWiz & IconQA & CLEVR & Flickr30k \\
    \midrule
       ImgNet-R & 91.23 &  &  &  &  &  \\
       ArxivQA & 90.07 & 92.80 &  &  &  &  \\
       VizWiz & 80.30 & 92.03 & 60.22 &  &  &  \\
       IconQA & 80.53 & 77.20 & 51.19 & 53.73 &  &  \\
       CLEVR & 79.37 & 77.40 & 48.68 & 51.30 & 71.67 & \\
       Flickr30k & 70.07 & 77.70 & 44.69 & 50.03 & 54.03 & 57.34  \\
    \bottomrule
    \end{tabular}}
\end{subtable}\hfill
\begin{subtable}[t]{0.49\linewidth}
\centering
\resizebox{\linewidth}{!}{
    \begin{tabular}{lcccccc}
    \toprule
       \textbf{ModalPrompt} & ImgNet-R & ArxivQA & VizWiz & IconQA & CLEVR & Flickr30k \\
    \midrule
       ImgNet-R & 51.10 &  &  &  &  &   \\
       ArxivQA & 51.03 & 87.37 &  &  &  &  \\
       VizWiz & 51.10 & 87.37 & 48.14 &  &  &  \\
       IconQA & 51.20 & 87.27 & 48.03 & 39.30 &  &  \\
       CLEVR & 51.17 & 87.23 & 47.96 & 39.40 & 47.07 & \\
       Flickr30k & 51.07 & 87.27 & 48.11 & 39.23 & 46.57 & 42.93  \\
    \bottomrule
    \end{tabular}}
\end{subtable}
\begin{subtable}[t]{0.49\linewidth}
    \centering
    \resizebox{\linewidth}{!}{
    \begin{tabular}{lcccccc}
    \toprule
       \textbf{CL-MoE} & ImgNet-R & ArxivQA & VizWiz & IconQA & CLEVR & Flickr30k \\
    \midrule
       ImgNet-R & 91.30 &  &  &  &  &   \\
       ArxivQA & 87.40 & 92.27 &  &  &  &  \\
       VizWiz & 76.23 & 90.53 & 60.18 &  &  &  \\
       IconQA & 77.33 & 76.20 & 51.09 & 54.33 &  &  \\
       CLEVR & 75.73 & 75.70 & 48.27 & 52.80 & 73.23 & \\
       Flickr30k & 66.33 & 77.00 & 44.78 & 51.87 & 53.53 & 57.42  \\
    \bottomrule
    \end{tabular}}
\end{subtable}\hfill
\begin{subtable}[t]{0.49\linewidth}
\centering
\resizebox{\linewidth}{!}{
    \begin{tabular}{lcccccc}
    \toprule
       \textbf{HiDe} & ImgNet-R & ArxivQA & VizWiz & IconQA & CLEVR & Flickr30k \\
    \midrule
       ImgNet-R & 90.87 &  &  &  &  &   \\
       ArxivQA & 88.83 & 92.27 &  &  &  &  \\
       VizWiz & 88.17 & 91.90 & 54.46 &  &  &  \\
       IconQA & 87.00 & 89.70 & 46.51 & 73.13 &  &  \\
       CLEVR & 86.23 & 90.07 & 44.28 & 64.37 & 54.77 & \\
       Flickr30k & 84.03 & 90.73 & 44.43 & 58.93 & 41.37 & 54.25  \\
    \bottomrule
    \end{tabular}}
\end{subtable}
\begin{subtable}[t]{0.49\linewidth}
    \centering
    \resizebox{\linewidth}{!}{
    \begin{tabular}{lcccccc}
    \toprule
       \textbf{SEFE} & ImgNet-R & ArxivQA & VizWiz & IconQA & CLEVR & Flickr30k \\
    \midrule
       ImgNet-R & 90.80 &  &  &  &  &   \\
       ArxivQA & 90.23 & 91.23 &  &  &  &  \\
       VizWiz & 83.60 & 89.73 & 61.24 &  &  &  \\
       IconQA & 84.17 & 77.20 & 56.36 & 78.93 &  &  \\
       CLEVR & 82.90 & 77.13 & 53.54 & 71.43 & 75.73 & \\
       Flickr30k & 80.83 & 78.00 & 47.01 & 69.63 & 65.83 & 57.92  \\
    \bottomrule
    \end{tabular}}
\end{subtable}\hfill
\begin{subtable}[t]{0.49\linewidth}
\centering
\resizebox{\linewidth}{!}{
    \begin{tabular}{lcccccc}
    \toprule
       \textbf{DISCO} & ImgNet-R & ArxivQA & VizWiz & IconQA & CLEVR & Flickr30k \\
    \midrule
       ImgNet-R & 90.97 &  &  &  &  &   \\
       ArxivQA & 90.07 & 91.77 &  &  &  &  \\
       VizWiz & 89.40 & 92.53 & 61.39 &  &  &  \\
       IconQA & 88.50 & 93.97 & 58.61 & 78.00 &  &  \\
       CLEVR & 88.03 & 93.40 & 57.11 & 70.93 & 76.40 & \\
       Flickr30k & 87.43 & 93.07 & 46.96 & 68.13 & 65.70 & 56.69  \\
    \bottomrule
    \end{tabular}}
\end{subtable}
\end{table*}

\begin{table*}[t]
\centering
\caption{Result matrices for different methods on the UCIT benchmark and InternVL.}
\label{tab:table5}
\begin{subtable}[t]{0.49\linewidth}
\centering
\resizebox{\linewidth}{!}{
\begin{tabular}{lcccccc}
\toprule
\textbf{LoRA-FT} & ImgNet-R & ArxivQA & VizWiz & IconQA & CLEVR & Flickr30k \\
\midrule
ImgNet-R & 95.40 & & & & & \\
ArxivQA & 94.40 & 92.70 & & & & \\
VizWiz & 87.50 & 92.73 & 62.53 & & & \\
IconQA & 86.10 & 78.63 & 54.95 & 82.87 & & \\
CLEVR & 83.27 & 78.33 & 49.92 & 71.07 & 82.87 & \\
Flickr30k & 76.50 & 78.60 & 45.57 & 71.13 & 61.63 & 58.82 \\
\bottomrule
\end{tabular}}
\end{subtable}\hfill
\begin{subtable}[t]{0.49\linewidth}
\centering
\resizebox{\linewidth}{!}{
\begin{tabular}{lcccccc}
\toprule
\textbf{O-LoRA} & ImgNet-R & ArxivQA & VizWiz & IconQA & CLEVR & Flickr30k \\
\midrule
ImgNet-R & 94.57 & & & & & \\
ArxivQA & 94.30 & 94.93 & & & & \\
VizWiz & 92.30 & 94.30 & 60.77 & & & \\
IconQA & 91.37 & 92.90 & 50.81 & 71.60 & & \\
CLEVR & 91.23 & 93.73 & 45.70 & 67.00 & 47.87 & \\
Flickr30k & 90.30 & 93.53 & 46.26 & 68.40 & 65.17 & 56.06 \\
\bottomrule
\end{tabular}}
\end{subtable}
\begin{subtable}[t]{0.49\linewidth}
    \centering
    \resizebox{\linewidth}{!}{
    \begin{tabular}{lcccccc}
    \toprule
        \textbf{MoELoRA} & ImgNet-R & ArxivQA & VizWiz & IconQA & CLEVR & Flickr30k \\
    \midrule
    ImgNet-R & 95.00 & & & & & \\
    ArxivQA & 93.57 & 93.73 & & & & \\
    VizWiz & 84.57 & 93.63 & 62.68 & & & \\
    IconQA & 81.43 & 79.30 & 57.85 & 82.23 & & \\
    CLEVR & 76.57 & 79.63 & 49.57 & 68.93 & 83.30 & \\
    Flickr30k & 68.57 & 80.47 & 45.43 & 69.23 & 65.93 & 58.57 \\
    \bottomrule
    \end{tabular}}
\end{subtable}\hfill
\begin{subtable}[t]{0.49\linewidth}
\centering
\resizebox{\linewidth}{!}{
    \begin{tabular}{lcccccc}
    \toprule
       \textbf{ModalPrompt} & ImgNet-R & ArxivQA & VizWiz & IconQA & CLEVR & Flickr30k \\
    \midrule
    ImgNet-R & 57.37 & & & & & \\
    ArxivQA & 57.40 & 63.87 & & & & \\
    VizWiz & 57.27 & 63.83 & 42.14 & & & \\
    IconQA & 57.30 & 63.87 & 42.03 & 37.20 & & \\
    CLEVR & 57.30 & 63.97 & 41.96 & 37.33 & 47.53 & \\
    Flickr30k & 57.10 & 63.90 & 42.11 & 37.40 & 47.50 & 45.76 \\
    \bottomrule
    \end{tabular}}
\end{subtable}
\begin{subtable}[t]{0.49\linewidth}
    \centering
    \resizebox{\linewidth}{!}{
    \begin{tabular}{lcccccc}
    \toprule
       \textbf{CL-MoE} & ImgNet-R & ArxivQA & VizWiz & IconQA & CLEVR & Flickr30k \\
    \midrule
    ImgNet-R & 94.63 & & & & & \\
    ArxivQA & 93.93 & 93.57 & & & & \\
    VizWiz & 85.37 & 91.73 & 61.60 & & & \\
    IconQA & 82.70 & 78.47 & 52.80 & 57.40 & & \\
    CLEVR & 80.43 & 78.47 & 48.24 & 55.23 & 62.93 & \\
    Flickr30k & 74.23 & 78.77 & 44.77 & 53.90 & 72.73 & 58.23 \\
    \bottomrule
    \end{tabular}}
\end{subtable}\hfill
\begin{subtable}[t]{0.49\linewidth}
\centering
\resizebox{\linewidth}{!}{
    \begin{tabular}{lcccccc}
    \toprule
       \textbf{HiDe} & ImgNet-R & ArxivQA & VizWiz & IconQA & CLEVR & Flickr30k \\
    \midrule
    ImgNet-R & 95.20 & & & & & \\
    ArxivQA & 94.90 & 93.73 & & & & \\
    VizWiz & 93.97 & 93.67 & 59.70 & & & \\
    IconQA & 92.33 & 93.57 & 52.43 & 80.33 & & \\
    CLEVR & 91.70 & 93.80 & 48.39 & 72.17 & 73.70 & \\
    Flickr30k & 90.00 & 93.77 & 50.49 & 71.03 & 61.10 & 55.08 \\
    \bottomrule
    \end{tabular}}
\end{subtable}
\begin{subtable}[t]{0.49\linewidth}
    \centering
    \resizebox{\linewidth}{!}{
    \begin{tabular}{lcccccc}
    \toprule
       \textbf{SEFE} & ImgNet-R & ArxivQA & VizWiz & IconQA & CLEVR & Flickr30k \\
    \midrule
    ImgNet-R & 94.60 & & & & & \\
    ArxivQA & 94.10 & 92.97 & & & & \\
    VizWiz & 90.47 & 90.10 & 61.37 & & & \\
    IconQA & 88.33 & 72.50 & 56.01 & 74.50 & & \\
    CLEVR & 87.50 & 75.97 & 49.66 & 67.80 & 59.67 & \\
    Flickr30k & 85.93 & 76.90 & 47.42 & 69.77 & 63.37 & 58.19 \\
    \bottomrule
    \end{tabular}}
\end{subtable}\hfill
\begin{subtable}[t]{0.49\linewidth}
\centering
\resizebox{\linewidth}{!}{
    \begin{tabular}{lcccccc}
    \toprule
       \textbf{DISCO} & ImgNet-R & ArxivQA & VizWiz & IconQA & CLEVR & Flickr30k \\
    \midrule
    ImgNet-R & 95.37 & & & & & \\
    ArxivQA & 94.43 & 92.43 & & & & \\
    VizWiz & 94.07 & 93.33 & 61.99 & & & \\
    IconQA & 92.80 & 93.80 & 58.35 & 83.07 & & \\
    CLEVR & 92.20 & 93.90 & 52.89 & 74.50 & 82.60 & \\
    Flickr30k & 92.13 & 94.13 & 48.11 & 73.90 & 67.53 & 58.06 \\
    \bottomrule
    \end{tabular}}
\end{subtable}
\end{table*}

\begin{table*}[t]
\centering
\caption{Result matrices for different methods on the DCL benchmark and LLaVA-1.5.}
\label{tab:table6}
\begin{subtable}[t]{0.45\linewidth}
\centering
\renewcommand{\arraystretch}{0.8} 
\resizebox{\linewidth}{!}{
\begin{tabular}{lcccccc}
\toprule
\textbf{LoRA-FT} & RS & Med & AD & Sci & Fin \\
\midrule
RS & 78.32 & & & & \\
Med & 74.68 & 57.53 & & & \\
AD & 68.93 & 47.19 & 52.15 & & \\
Sci & 75.12 & 45.56 & 38.46 & 49.44 & \\
Fin & 69.65 & 41.59 & 25.43 & 40.88 & 87.45 \\
\bottomrule
\end{tabular}}
\end{subtable}\hfill
\begin{subtable}[t]{0.45\linewidth}
\centering
\renewcommand{\arraystretch}{0.8} 
\resizebox{\linewidth}{!}{
\begin{tabular}{lcccccc}
\toprule
\textbf{O-LoRA} & RS & Med & AD & Sci & Fin \\
\midrule
RS & 79.25 & & & & \\
Med & 74.05 & 56.52 & & & \\
AD & 76.06 & 43.71 & 52.32 & & \\
Sci & 76.60 & 44.87 & 40.57 & 50.58 & \\
Fin & 74.64 & 44.42 & 30.02 & 41.47 & 87.15 \\
\bottomrule
\end{tabular}}
\end{subtable}
\begin{subtable}[t]{0.45\linewidth}
    \centering
    \renewcommand{\arraystretch}{0.8} 
    \resizebox{\linewidth}{!}{
    \begin{tabular}{lcccccc}
    \toprule
        \textbf{MoELoRA} & RS & Med & AD & Sci & Fin \\
    \midrule
    RS & 79.09 & & & & \\
    Med & 74.78 & 58.73 & & & \\
    AD & 77.69 & 43.72 & 51.47 & & \\
    Sci & 76.87 & 43.79 & 32.81 & 48.67 & \\
    Fin & 77.54 & 41.85 & 27.62 & 40.13 & 86.75 \\
    \bottomrule
    \end{tabular}}
\end{subtable}\hfill
\begin{subtable}[t]{0.45\linewidth}
\centering
\renewcommand{\arraystretch}{0.8} 
\resizebox{\linewidth}{!}{
    \begin{tabular}{lcccccc}
    \toprule
       \textbf{ModalPrompt} & RS & Med & AD & Sci & Fin \\
    \midrule
    RS & 53.12 & & & & \\
    Med & 53.02 & 45.82 & & & \\
    AD & 53.14 & 45.77 & 40.78 & & \\
    Sci & 53.03 & 45.65 & 40.78 & 41.83 & \\
    Fin & 53.63 & 45.68 & 40.77 & 41.81 & 87.82 \\
    \bottomrule
    \end{tabular}}
\end{subtable}
\begin{subtable}[t]{0.45\linewidth}
    \centering
    \renewcommand{\arraystretch}{0.8} 
    \resizebox{\linewidth}{!}{
    \begin{tabular}{lcccccc}
    \toprule
       \textbf{CL-MoE} & RS & Med & AD & Sci & Fin \\
    \midrule
    RS & 79.08 & & & & \\
    Med & 73.48 & 60.56 & & & \\
    AD & 72.61 & 44.42 & 51.62 & & \\
    Sci & 71.02 & 48.04 & 37.70 & 50.28 & \\
    Fin & 71.34 & 46.84 & 26.33 & 41.17 & 88.74 \\
    \bottomrule
    \end{tabular}}
\end{subtable}\hfill
\begin{subtable}[t]{0.45\linewidth}
\centering
\renewcommand{\arraystretch}{0.8} 
\resizebox{\linewidth}{!}{
    \begin{tabular}{lcccccc}
    \toprule
       \textbf{HiDe} & RS & Med & AD & Sci & Fin \\
    \midrule
    RS & 78.14 & & & & \\
    Med & 74.26 & 58.05 & & & \\
    AD & 74.90 & 42.94 & 39.65 & & \\
    Sci & 75.43 & 44.91 & 38.33 & 46.44 & \\
    Fin & 74.31 & 48.95 & 33.21 & 38.54 & 81.55 \\
    \bottomrule
    \end{tabular}}
\end{subtable}
\begin{subtable}[t]{0.45\linewidth}
    \centering
    \renewcommand{\arraystretch}{0.8} 
    \resizebox{\linewidth}{!}{
    \begin{tabular}{lcccccc}
    \toprule
       \textbf{SEFE} & RS & Med & AD & Sci & Fin \\
    \midrule
    RS & 78.27 & & & & \\
    Med & 76.32 & 58.42 & & & \\
    AD & 77.22 & 49.13 & 52.49 & & \\
    Sci & 77.83 & 47.70 & 43.01 & 49.04 & \\
    Fin & 77.26 & 50.37 & 37.21 & 40.87 & 86.82 \\
    \bottomrule
    \end{tabular}}
\end{subtable}\hfill
\begin{subtable}[t]{0.45\linewidth}
\centering
\renewcommand{\arraystretch}{0.8} 
\resizebox{\linewidth}{!}{
    \begin{tabular}{lcccccc}
    \toprule
       \textbf{DISCO} & RS & Med & AD & Sci & Fin \\
    \midrule
    RS & 79.00 & & & & \\
    Med & 77.35 & 52.77 & & & \\
    AD & 77.60 & 50.01 & 52.22 & & \\
    Sci & 76.63 & 44.90 & 49.04 & 50.69 & \\
    Fin & 76.49 & 44.48 & 44.84 & 46.61 & 89.22 \\
    \bottomrule
    \end{tabular}}
\end{subtable}
\end{table*}

\begin{table*}[t]
\centering
\caption{Result matrices for different methods on the DCL benchmark and InternVL.}
\label{tab:table7}
\begin{subtable}[t]{0.45\linewidth}
\centering
\renewcommand{\arraystretch}{0.8} 
\resizebox{\linewidth}{!}{
\begin{tabular}{lcccccc}
\toprule
\textbf{LoRA-FT} & RS & Med & AD & Sci & Fin \\
\midrule
RS & 81.29 & & & & \\
Med & 75.71 & 65.92 & & & \\
AD & 69.38 & 56.87 & 53.56 & & \\
Sci & 71.12 & 53.75 & 46.83 & 53.48 & \\
Fin & 69.93 & 52.17 & 33.04 & 42.67 & 91.07 \\
\bottomrule
\end{tabular}}
\end{subtable}\hfill
\begin{subtable}[t]{0.45\linewidth}
\centering
\renewcommand{\arraystretch}{0.8} 
\resizebox{\linewidth}{!}{
\begin{tabular}{lcccccc}
\toprule
\textbf{O-LoRA} & RS & Med & AD & Sci & Fin \\
\midrule
RS & 81.44 & & & & \\
Med & 79.98 & 64.78 & & & \\
AD & 77.75 & 57.91 & 42.41 & & \\
Sci & 76.53 & 53.90 & 38.92 & 50.40 & \\
Fin & 74.48 & 54.16 & 39.60 & 48.30 & 88.54 \\
\bottomrule
\end{tabular}}
\end{subtable}
\begin{subtable}[t]{0.45\linewidth}
    \centering
    \renewcommand{\arraystretch}{0.8} 
    \resizebox{\linewidth}{!}{
    \begin{tabular}{lcccccc}
    \toprule
        \textbf{MoELoRA} & RS & Med & AD & Sci & Fin \\
    \midrule
    RS & 81.22 & & & & \\
    Med & 77.56 & 66.00 & & & \\
    AD & 74.56 & 58.74 & 53.62 & & \\
    Sci & 72.62 & 54.77 & 47.65 & 52.75 & \\
    Fin & 69.90 & 52.08 & 33.17 & 42.19 & 90.58 \\
    \bottomrule
    \end{tabular}}
\end{subtable}\hfill
\begin{subtable}[t]{0.45\linewidth}
\centering
\renewcommand{\arraystretch}{0.8} 
\resizebox{\linewidth}{!}{
    \begin{tabular}{lcccccc}
    \toprule
       \textbf{ModalPrompt} & RS & Med & AD & Sci & Fin \\
    \midrule
    RS & 50.16 & & & & \\
    Med & 50.02 & 31.00 & & & \\
    AD & 49.97 & 30.92 & 14.07 & & \\
    Sci & 49.95 & 30.90 & 14.09 & 33.99 & \\
    Fin & 50.20 & 30.95 & 14.11 & 33.91 & 65.45 \\
    \bottomrule
    \end{tabular}}
\end{subtable}
\begin{subtable}[t]{0.45\linewidth}
    \centering
    \renewcommand{\arraystretch}{0.8} 
    \resizebox{\linewidth}{!}{
    \begin{tabular}{lcccccc}
    \toprule
       \textbf{CL-MoE} & RS & Med & AD & Sci & Fin \\
    \midrule
    RS & 81.55 & & & & \\
    Med & 78.51 & 66.85 & & & \\
    AD & 78.41 & 61.00 & 53.54 & & \\
    Sci & 78.44 & 57.37 & 49.07 & 52.94 & \\
    Fin & 78.12 & 52.51 & 35.53 & 42.69 & 91.24 \\
    \bottomrule
    \end{tabular}}
\end{subtable}\hfill
\begin{subtable}[t]{0.45\linewidth}
\centering
\renewcommand{\arraystretch}{0.8} 
\resizebox{\linewidth}{!}{
    \begin{tabular}{lcccccc}
    \toprule
       \textbf{HiDe} & RS & Med & AD & Sci & Fin \\
    \midrule
    RS & 81.24 & & & & \\
    Med & 79.59 & 64.71 & & & \\
    AD & 78.85 & 58.37 & 41.75 & & \\
    Sci & 78.33 & 58.51 & 39.94 & 49.99 & \\
    Fin & 75.40 & 57.66 & 36.73 & 41.48 & 88.59 \\
    \bottomrule
    \end{tabular}}
\end{subtable}
\begin{subtable}[t]{0.45\linewidth}
    \centering
    \renewcommand{\arraystretch}{0.8} 
    \resizebox{\linewidth}{!}{
    \begin{tabular}{lcccccc}
    \toprule
       \textbf{SEFE} & RS & Med & AD & Sci & Fin \\
    \midrule
    RS & 80.43 & & & & \\
    Med & 78.01 & 67.62 & & & \\
    AD & 79.04 & 62.36 & 54.46 & & \\
    Sci & 78.48 & 58.15 & 53.53 & 53.88 & \\
    Fin & 78.21 & 57.59 & 51.45 & 44.65 & 91.37 \\
    \bottomrule
    \end{tabular}}
\end{subtable}\hfill
\begin{subtable}[t]{0.45\linewidth}
\centering
\renewcommand{\arraystretch}{0.8} 
\resizebox{\linewidth}{!}{
    \begin{tabular}{lcccccc}
    \toprule
       \textbf{DISCO} & RS & Med & AD & Sci & Fin \\
    \midrule
    RS & 81.49 & & & & \\
    Med & 80.14 & 63.05 & & & \\
    AD & 78.87 & 57.42 & 53.77 & & \\
    Sci & 78.67 & 52.80 & 53.56 & 53.52 & \\
    Fin & 77.90 & 47.50 & 49.13 & 49.37 & 90.92 \\
    \bottomrule
    \end{tabular}}
\end{subtable}
\end{table*}

\begin{table*}[t]
\centering
\caption{Result matrices for different methods on the ACL benchmark and LLaVA-1.5.}
\label{tab:table8}
\begin{subtable}[t]{0.45\linewidth}
\centering
\renewcommand{\arraystretch}{0.8} 
\resizebox{\linewidth}{!}{
\begin{tabular}{lcccccc}
\toprule
\textbf{LoRA-FT} & OCR & Math & VP & APP \\
\midrule
OCR & 33.30 & & & \\
Math & 32.60 & 34.20 & & \\
VP & 31.70 & 32.80 & 65.10 & \\
APP & 23.60 & 33.70 & 55.84 & 32.50 \\
\bottomrule
\end{tabular}}
\end{subtable}\hfill
\begin{subtable}[t]{0.45\linewidth}
\centering
\renewcommand{\arraystretch}{0.8} 
\resizebox{\linewidth}{!}{
\begin{tabular}{lcccccc}
\toprule
\textbf{O-LoRA} & OCR & Math & VP & APP \\
\midrule
OCR & 32.90 & & & \\
Math & 29.80 & 33.60 & & \\
VP & 27.40 & 33.70 & 58.63 & \\
APP & 29.60 & 32.90 & 52.41 & 33.75 \\
\bottomrule
\end{tabular}}
\end{subtable}
\begin{subtable}[t]{0.45\linewidth}
    \centering
    \renewcommand{\arraystretch}{0.8} 
    \resizebox{\linewidth}{!}{
    \begin{tabular}{lcccccc}
    \toprule
        \textbf{MoELoRA} & OCR & Math & VP & APP \\
    \midrule
    OCR & 32.70 & & & \\
    Math & 32.50 & 33.30 & & \\
    VP & 30.80 & 33.00 & 64.59 & \\
    APP & 26.70 & 32.80 & 56.85 & 27.22 \\
    \bottomrule
    \end{tabular}}
\end{subtable}\hfill
\begin{subtable}[t]{0.45\linewidth}
\centering
\renewcommand{\arraystretch}{0.8} 
\resizebox{\linewidth}{!}{
    \begin{tabular}{lcccccc}
    \toprule
       \textbf{ModalPrompt} & OCR & Math & VP & APP \\
    \midrule
    OCR & 31.30 & & & \\
    Math & 30.80 & 30.20 & & \\
    VP & 28.10 & 32.40 & 62.94 & \\
    APP & 31.80 & 32.50 & 60.53 & 10.00 \\
    \bottomrule
    \end{tabular}}
\end{subtable}
\begin{subtable}[t]{0.45\linewidth}
    \centering
    \renewcommand{\arraystretch}{0.8} 
    \resizebox{\linewidth}{!}{
    \begin{tabular}{lcccccc}
    \toprule
       \textbf{CL-MoE} & OCR & Math & VP & APP \\
    \midrule
    OCR & 33.00 & & & \\
    Math & 32.30 & 33.60 & & \\
    VP & 30.20 & 32.50 & 64.72 & \\
    APP & 19.90 & 32.70 & 53.43 & 30.69 \\
    \bottomrule
    \end{tabular}}
\end{subtable}\hfill
\begin{subtable}[t]{0.45\linewidth}
\centering
\renewcommand{\arraystretch}{0.8} 
\resizebox{\linewidth}{!}{
    \begin{tabular}{lcccccc}
    \toprule
       \textbf{HiDe} & OCR & Math & VP & APP \\
    \midrule
    OCR & 33.40 & & & \\
    Math & 30.90 & 32.80 & & \\
    VP & 30.40 & 33.30 & 56.98 & \\
    APP & 24.60 & 32.10 & 46.32 & 28.75 \\
    \bottomrule
    \end{tabular}}
\end{subtable}
\begin{subtable}[t]{0.45\linewidth}
    \centering
    \renewcommand{\arraystretch}{0.8} 
    \resizebox{\linewidth}{!}{
    \begin{tabular}{lcccccc}
    \toprule
       \textbf{SEFE} & OCR & Math & VP & APP \\
    \midrule
    OCR & 33.00 & & & \\
    Math & 32.20 & 32.60 & & \\
    VP & 31.80 & 33.30 & 64.59 & \\
    APP & 26.00 & 33.40 & 57.74 & 33.75 \\
    \bottomrule
    \end{tabular}}
\end{subtable}\hfill
\begin{subtable}[t]{0.45\linewidth}
\centering
\renewcommand{\arraystretch}{0.8} 
\resizebox{\linewidth}{!}{
    \begin{tabular}{lcccccc}
    \toprule
       \textbf{DISCO} & OCR & Math & VP & APP \\
    \midrule
    OCR & 32.90 & & & \\
    Math & 31.80 & 33.40 & & \\
    VP & 31.00 & 34.50 & 59.64 & \\
    APP & 32.90 & 33.10 & 60.15 & 30.14 \\
    \bottomrule
    \end{tabular}}
\end{subtable}
\end{table*}

\begin{table*}[t]
\centering
\caption{Result matrices for different methods on the ACL benchmark and InternVL.}
\label{tab:table9}
\begin{subtable}[t]{0.45\linewidth}
\centering
\renewcommand{\arraystretch}{0.8} 
\resizebox{\linewidth}{!}{
\begin{tabular}{lcccccc}
\toprule
\textbf{LoRA-FT} & OCR & Math & VP & APP \\
\midrule
OCR & 32.20 & & & \\
Math & 33.10 & 33.30 & & \\
VP & 31.80 & 32.30 & 68.02 & \\
APP & 21.40 & 32.80 & 60.28 & 29.86 \\
\bottomrule
\end{tabular}}
\end{subtable}\hfill
\begin{subtable}[t]{0.45\linewidth}
\centering
\renewcommand{\arraystretch}{0.8} 
\resizebox{\linewidth}{!}{
\begin{tabular}{lcccccc}
\toprule
\textbf{O-LoRA} & OCR & Math & VP & APP \\
\midrule
OCR & 32.70 & & & \\
Math & 31.10 & 34.20 & & \\
VP & 30.20 & 33.00 & 63.20 & \\
APP & 25.50 & 32.30 & 64.59 & 24.44 \\
\bottomrule
\end{tabular}}
\end{subtable}
\begin{subtable}[t]{0.45\linewidth}
    \centering
    \renewcommand{\arraystretch}{0.8} 
    \resizebox{\linewidth}{!}{
    \begin{tabular}{lcccccc}
    \toprule
        \textbf{MoELoRA} & OCR & Math & VP & APP \\
    \midrule
    OCR & 32.20 & & & \\
    Math & 29.90 & 33.30 & & \\
    VP & 29.20 & 32.80 & 67.64 & \\
    APP & 17.20 & 32.70 & 55.33 & 32.50 \\
    \bottomrule
    \end{tabular}}
\end{subtable}\hfill
\begin{subtable}[t]{0.45\linewidth}
\centering
\renewcommand{\arraystretch}{0.8} 
\resizebox{\linewidth}{!}{
    \begin{tabular}{lcccccc}
    \toprule
       \textbf{ModalPrompt} & OCR & Math & VP & APP \\
    \midrule
    OCR & 30.40 & & & \\
    Math & 30.40 & 31.50 & & \\
    VP & 11.50 & 31.50 & 56.89 & \\
    APP & 25.90 & 31.60 & 55.58 & 11.39 \\
    \bottomrule
    \end{tabular}}
\end{subtable}
\begin{subtable}[t]{0.45\linewidth}
    \centering
    \renewcommand{\arraystretch}{0.8} 
    \resizebox{\linewidth}{!}{
    \begin{tabular}{lcccccc}
    \toprule
       \textbf{CL-MoE} & OCR & Math & VP & APP \\
    \midrule
    OCR & 31.80 & & & \\
    Math & 29.90 & 33.40 & & \\
    VP & 28.60 & 33.10 & 67.89 & \\
    APP & 25.10 & 31.90 & 60.91 & 27.50 \\
    \bottomrule
    \end{tabular}}
\end{subtable}\hfill
\begin{subtable}[t]{0.45\linewidth}
\centering
\renewcommand{\arraystretch}{0.8} 
\resizebox{\linewidth}{!}{
    \begin{tabular}{lcccccc}
    \toprule
       \textbf{HiDe} & OCR & Math & VP & APP \\
    \midrule
    OCR & 32.40 & & & \\
    Math & 29.70 & 33.60 & & \\
    VP & 28.60 & 33.00 & 61.80 & \\
    APP & 17.80 & 33.00 & 41.12 & 33.33 \\
    \bottomrule
    \end{tabular}}
\end{subtable}
\begin{subtable}[t]{0.45\linewidth}
    \centering
    \renewcommand{\arraystretch}{0.8} 
    \resizebox{\linewidth}{!}{
    \begin{tabular}{lcccccc}
    \toprule
       \textbf{SEFE} & OCR & Math & VP & APP \\
    \midrule
    OCR & 32.60 & & & \\
    Math & 30.90 & 33.00 & & \\
    VP & 30.90 & 33.70 & 68.27 & \\
    APP & 19.00 & 32.20 & 62.18 & 31.94 \\
    \bottomrule
    \end{tabular}}
\end{subtable}\hfill
\begin{subtable}[t]{0.45\linewidth}
\centering
\renewcommand{\arraystretch}{0.8} 
\resizebox{\linewidth}{!}{
    \begin{tabular}{lcccccc}
    \toprule
       \textbf{DISCO} & OCR & Math & VP & APP \\
    \midrule
    OCR & 31.90 & & & \\
    Math & 31.70 & 34.00 & & \\
    VP & 32.10 & 33.50 & 63.45 & \\
    APP & 30.60 & 33.10 & 65.36 & 27.50 \\
    \bottomrule
    \end{tabular}}
\end{subtable}
\end{table*}

\end{document}